\documentclass{article} 
\usepackage{iclr2024_workshop,times}

\usepackage{amsmath,amsfonts,bm}

\def\eqref#1{equation~\ref{#1}}

\def\1{\bm{1}}

\DeclareMathAlphabet{\mathsfit}{\encodingdefault}{\sfdefault}{m}{sl}
\SetMathAlphabet{\mathsfit}{bold}{\encodingdefault}{\sfdefault}{bx}{n}

\usepackage{hyperref}
\usepackage{url}
\usepackage{booktabs}
\usepackage{graphicx}
\usepackage{float}
\title{RBF-PINN: Non-Fourier Positional Embedding in Physics-Informed Neural Networks}

\author{Chengxi Zeng\thanks{Correspondence Author. The full version of the paper can be found in \url{arxiv.org/abs/2402.06955}. \\
The code can be found in repo \url{github.com/SimonZeng7108/RBF-PINN/tree/master}.}, Tilo Burghardt \& Alberto M.~Gambaruto  \\
University of Bristol, UK\\
\texttt{\{cz15306, tb2935, alberto.gambaruto\}@bristol.ac.uk} \\
}

\iclrfinalcopy 
\begin{document}

\maketitle

\begin{abstract}
While many recent Physics-Informed Neural Networks (PINNs) variants have had considerable success in solving Partial Differential Equations, the empirical benefits of feature mapping drawn from the broader Neural Representations research have been largely overlooked. We highlight the limitations of widely used Fourier-based feature mapping in certain situations and suggest the use of the conditionally positive definite Radial Basis Function. The empirical findings demonstrate the effectiveness of our approach across a variety of forward and inverse problem cases. Our method can be seamlessly integrated into coordinate-based input neural networks and contribute to the wider field of PINNs research.

\end{abstract}
\vspace{-0.2in}
\section{Introduction}
\vspace{-0.1in}
Many scientific phenomena can be described by sets of differential equations (DEs). The prior physics knowledge is then formulated as regularisers that can be embedded in modern machine learning algorithms supported by the Universal Approximation Theorem~\cite{Hornik1989MultilayerFN}. Physics-Informed Machine Learning (PIML)~\cite{Karniadakis2021PhysicsinformedML} is a learning paradigm that combines data-driven models with physical laws and domain knowledge to solve complex problems in science and engineering. It has recently attained remarkable achievements in a wide range of scientific research such as Electronics~\cite{Smith2022PhysicsInformedIR, Hu2023SyncTREE, Nicoli2023Quant}, Dynamical System~\cite{Thangamuthu2022Dynamical, Ni2022NTFieldsNT}, Meteorology~\cite{kashinath2021physics, Giladi2021PhysicsAwareDW} and Medical Image~\cite{Goyeneche2023ResoNet, Salehi2021PhysGNNAP, pokkunuru2023improved}. One of the leading methods is called Physics-Informed Neural Networks (PINNs)~\cite{Raissi2019PhysicsinformedNN}. By adding the DEs as penalty terms in the deep Neural Networks (NN), it exploits the differentiability of NN to compute derivatives of the explicit functions and introduces domain-specific regularisation during optimisation. Adhering to conventional solvers, the PINNs formulation necessitates the specification of initial/boundary conditions (IC/BC) within a confined spatial-temporal domain. Boundary sampling points and domain collocation points are used to evaluate the residuals of the conditions and DEs via an overparameterised NN. The objective is to optimise the NN by minimising the residuals, resulting in a converged parameter space. This parameter space can then serve as a surrogate model that accurately represents the solution space of the DEs. The PINNs can be formulated as follows:\\
\vspace{-0.1in}
\begin{equation}
\mathcal{D}[u(x, t; \alpha_i)] = F(x, t), t \in \mathcal{T}[0, T], \forall x \in \Omega \text{ and }
\mathcal{B}[u(x, t)] = H(x, t), t \in \mathcal{T}[0, T], x \in \partial\Omega .
\end{equation}
\vspace{-0.15in}\\
where $\mathcal{D}[\cdot]$ is the differential operator and  $\mathcal{B}[\cdot]$ is the boundary operator, $x$ and $t$ are the independent variables in spatial and temporal domains $\Omega$ and $\mathcal{T}$, respectively. The $\alpha_i$ are coefficients of the DE system and remain wholly or partially unknown in Inverse Problems. For time-dependent PDEs, the initial condition can be treated as a special type of BC. $F$ and $H$ are arbitrary functions.\\
PINNs are parameterised by $\theta$, the solution space represented can give numerical solution $\hat u_\theta$ at any $x$ and $t$ within the domain. And the training loss functions are defined as follows:\\
\vspace{-0.13in}
\begin{equation}
\label{equation: loss}
\mathcal{L}(\theta; \mathcal{X}_{(x, t)}) = \frac{\lambda_r}{N_r}\sum_{i=1}^{N_r} \left | \mathcal{D}[\hat u_\theta(x_{r}^i)] - F(x_r^i) \right |^2 + \frac{\lambda_{bc}}{N_{bc}}\sum_{i=1}^{N_{bc}} \left | \mathcal{B}[\hat u_\theta(x_{bc}^i)] - H(x_{bc}^i) \right |^2
\end{equation}
\vspace{-0.08in}\\
where $\{x_r^i\}_{i = 1}^{N_r}$ and $\{x_{bc}^i\}_{i = 1}^{N_{bc}}$ are domain collocation points and boundary points, they are evaluated by computing the mean squared error. $\lambda_r$ and $\lambda_{bc}$ are the corresponding weights of each term.\\
\newpage
\begin{figure*}[ht]
\begin{center}
\centerline{\includegraphics[height=3.2cm]{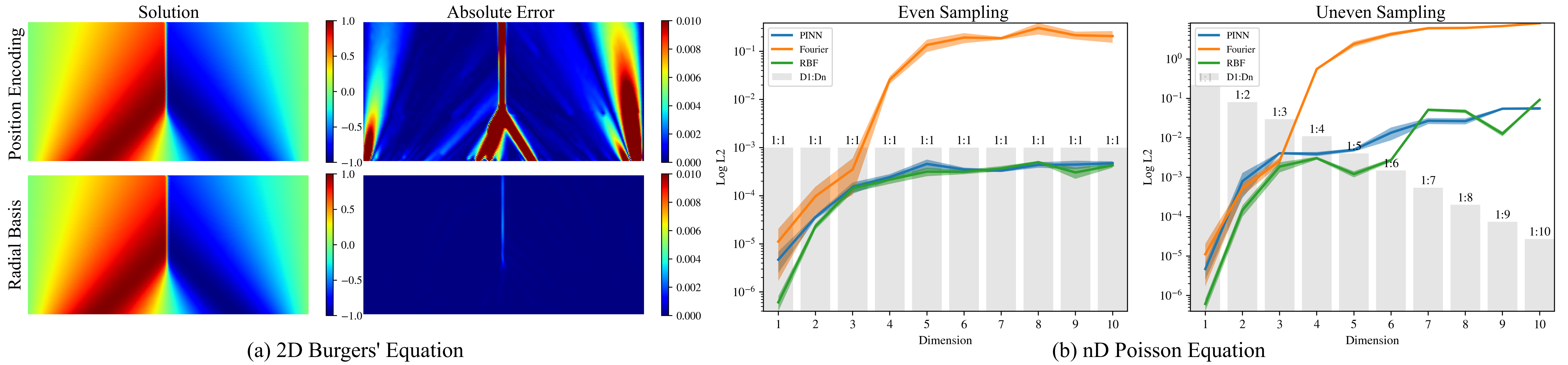}}
\vspace{-0.2in}
\caption{(a) Solutions using PINNs to solve Diffusion Equation with Positional Encoding (Top) and Our RBF (Bottom) feature mappings ; (b) $L2$ error on nD Poisson equation from 1 to 10 dimensions.}
\label{Fig: error demo}
\end{center}
\end{figure*}
\vspace{-0.3in}
Although significant advances have been achieved in prior research, there has been limited exploration of feature mapping, with only a few studies~\cite{WANG2021113938, Wong2022Sinusoidal} recognising its potential within PINNs. Originally introduced in Natural Language Processing (NLP), feature mapping aims to map the input to a higher-dimensional feature space for a better neural representation. It was subsequently identified as an effective approach for mitigating spectral bias in visual representations~\cite{Tancik2020FourierFL}. Feature mapping is a broader term for positional encoding that can involve either fixed encoding or trainable embedding. This simple one-layer projection can map the spatio-temporal input $\mathbf{x}$ to a higher dimension feature space, $\Phi: \mathbf{x} \in \mathbb{R}^n \rightarrow \mathbb{R}^m$, and typically $n \ll m$.\\
In PINNs, \citet{Wang2020WhenAW} has leveraged the `Neural Tangent Kernel' (NTK) theory that reveals PINNs suffer from `Spectral Bias' in the infinite-width limit. The NTK exhibits sensitivity to both input and network parameters. Its trait is particularly contingent on factors such as the input gradient of the model, the variance at each layer, and the nonlinear activations. Hence, the training dynamics of PINNs are significantly influenced by the input before parameterised layers. \\
Our contribution in this paper can be summarised as: First, we show the limitations and shortcomings of the widely used Fourier-based feature mappings in some Partial Differential Equations (PDEs) and thoroughly benchmark a wide range of feature mapping methods, including some which have not been employed in PINNs before. Secondly, we present a framework for designing feature mapping functions and introduce a conditional positive definite Radial Basis Function. This method surpasses Fourier-based feature mapping in various forward and inverse tasks.\\
\vspace{-0.2in}
\section{Limitation of Fourier Features}
\vspace{-0.1in}
Basic Fourier features can lead to undesired artifacts, as illustrated in Figure~\ref{Fig: error demo}(a) (detailed equations can be found in Appendix~\ref{Append: Benchmark}). When applying Positional Encoding to address Burgers' equations, a notable prediction error is observed in the area approaching a jump solution. This phenomenon is akin to the Gibbs phenomenon in signal processing, where the approximated function value by a finite number of terms in its Fourier series tends to overshoot and oscillate around a discontinuity.\\
Another unexpected experimental result that shows poor performance is observed when utilising Random Fourier Features~\cite{Tancik2020FourierFL} in high-dimensional problems, as depicted in Figure~\ref{Fig: error demo}(b). Two cases are set up, one case with each dimension of evenly sampled points and another with uneven sampling points, where the number of sampling points $x_r$ is set as $\frac{1}{D}$. The latter example resembles the unsteady Navier-Stokes equations for fluid dynamics, which have dense samples in the spatial domains, but sparse sampling in the temporal dimension. Despite hyperparameter tuning for the Fourier feature, including the arbitrary scale $\sigma$ and the number of Fourier features, none of the attempts reduce the elevated error in high-dimensional cases. The experiments are repeated with 3 random seeds, and standard deviations are displayed in the highlighted area. 

\section{Proposed Method}
\vspace{-0.1in}
In NTK theory, the multi-layer perceptions (MLPs) function is approximated by the convolution of the stationary composed NTK function $K_{\text{COMP}}=K_{NTK}\circ K_\Phi$ with weighted Dirac delta over the input $\mathbf{x}$ (background in Spectral Bias and Composed NTK are in Appendix~\ref{Append: Spectral}), we can formulate the $K_{\text{COMP}}$ by:
\begin{equation}
\begin{aligned}
K_{\text{COMP}}(\mathbf{x}) &= (K_{\text{COMP}}*\delta_x)(\mathbf{x})\\
&= \int K_{\text{COMP}}(\mathbf{x}')\delta(\mathbf{x}-\mathbf{x}')d\mathbf{x}\\
&\approx \int K_{\text{COMP}}(\mathbf{x}')K_\Phi(\mathbf{x}-\mathbf{x}')d\mathbf{x}
\end{aligned}
\end{equation}
The accuracy of the continuous approximation can be analysed by Taylor series expansion:
\begin{equation}
\label{equa: Taylor}
\begin{aligned}
K_{\text{COMP}}(\mathbf{x}) &= \int(K_{\text{COMP}}(\mathbf{x}) + \nabla_{\mathbf{x}}K_{\text{COMP}}(\mathbf{x}-\mathbf{x}') \\
&+ \frac{1}{2}(\mathbf{x}-\mathbf{x}')\nabla^2 K_{\text{COMP}}(\mathbf{x}-\mathbf{x}')\\
&+ \mathcal{O}((\mathbf{x}-\mathbf{x}')^3))K_\Phi(\mathbf{x}-\mathbf{x}') d\mathbf{x}\\
&= K_{\text{COMP}}(\mathbf{\mathbf{x}})\int K_\Phi(\mathbf{x}-\mathbf{x}') d\mathbf{x}  \\
&+\nabla_{\mathbf{x}}K_{\text{COMP}}(\mathbf{x}-\mathbf{x}')\int(\mathbf{x}-\mathbf{x}') K_\Phi(\mathbf{x}-\mathbf{x}') d\mathbf{x} \\
&+\mathcal{O}((\mathbf{x}-\mathbf{x}')^2)
\end{aligned}
\end{equation}
Ensuring the first-order accuracy of the composing kernel requires the term $\int K_\Phi(\mathbf{x}-\mathbf{x}') d\mathbf{x} = 1$, and the second term in Equation~\ref{equa: Taylor} must be 0. This can be accomplished by normalising the feature mapping function and ensuring that it satisfies a symmetry condition.
We propose a positive definite Radial Basis Function (RBF) for such a kernel, and its formulation is given by:
\begin{equation}
\Phi(\mathbf{x}) = \frac{\sum^m_iw_i\varphi(|\mathbf{x}-c_i|)}{\sum^m_i\varphi(|\mathbf{x}-c_i|)}
\end{equation}
where $\mathbf{x}\in R^n$ is the input data, $\mathbf{c} \in R^{n\times m}$ are the centres of the RBFs and are trainable parameters and $w$ is the weight matrix for the feature mapping layer. 
A natural choice for the RBF can be the Gaussian function, $\varphi(x)=e^{-\frac{|\mathbf{x}-\mathbf{c}|^2}{\mathbf{\sigma}^2}}$, where $\mathbf{\sigma}$ is a random initialised trainable parameter. If we choose the same number of features as the input size (i.e. $n=m$), this method provides an approximate computation of the desired function value through kernel regression. Unfortunately, in the context of PINNs, the training input size is typically large, making it impractical to scale in this manner. Through empirical study, we demonstrate that a few hundred RBFs prove sufficient to outperform other types of feature mapping functions. During initialisation, $\mathbf{c}$ is sampled from a standard Gaussian distribution.\\
\subsection{Conditionally positive definite RBF}
In the infinite-width limit, each layer of the Neural Network is treated as a linear system. To guarantee a unique solution, one approach involves introducing conditionally positive definite radial functions by incorporating polynomial terms. The weights serve as Lagrange multipliers, enabling constraints on the RBF coefficients in the parameter space~\cite{Farazandeh2021ARR}. We denote this method as RBF-P throughout the paper. Therefore, the feature mapping function is adjusted to:
\begin{equation}
\Phi(\mathbf{x}) = \frac{\sum^m_iw^m_i\varphi(|\mathbf{x}-c_i|)}{\sum^m_i\varphi(|\mathbf{x}-c_i|)} + \sum^k_jw^k_j P(\mathbf{x})
\end{equation}
Where P is the polynomial function.  
In the feature mapping layer, it can be represented as:
\begin{equation}
\begin{bmatrix} 
f_1 \\
\vdots \\
f_N \\ 
\end{bmatrix} = 
\begin{bmatrix} 
\varphi(r_1^1) & \hdots & \varphi(r_1^m) & \mid & 1 & x_1 & x^{k} \\
\vdots & \ddots & \vdots& |  & \vdots & \vdots & \vdots \\
\varphi(r_N^1) & \hdots & \varphi(r_N^m) & \mid  & 1 & x_N & x_N^{k} \\
\end{bmatrix}
\begin{bmatrix} 
W^m\\
- \\
W^k \\
\end{bmatrix}
\end{equation}
where $r = x-c$ and $P$ is the order of the polynomial term.\\
Based on empirical findings, we observe that the polynomial term is highly effective in nonlinear function approximation, particularly in equations like the Burgers Equation and Navier-Stokes Equation, all while incurring minimal computational overhead. \\
By this principle, we can use many other types of RBF without too many restrictions. Other types of RBF are shown in Appendix~\ref{appendix: ablation study} Table~\ref{tab: RBF types}.

\section{Empirical Results}

\begin{table*}[h]
\centering
\small
\caption{$L2$ error on varies of PDEs with different feature mapping. The best results are in\colorbox{blue!25}{Blue}. Full experimental results with standard deviations are shown in Appendix~\ref{Append: full results}.}
\label{tab: forward pdes}
\begin{tabular}{@{}cccccccccc@{}}
\toprule
                               & PINN                           & BE                 & PE            & FF                & SF             & CT                     & CG                       & \textbf{RBF}                            & \textbf{RBF-P}   \\ \midrule
\multicolumn{1}{l|}{Wave}      & \multicolumn{1}{l|}{3.73e-1} & \multicolumn{1}{l|}{1.04e0}  & \multicolumn{1}{l|}{1.01e0}  & \multicolumn{1}{l|}{\colorbox{blue!25}{2.38e-3}} & \multicolumn{1}{l|}{7.93e-3} & \multicolumn{1}{l|}{1.11e0}  & \multicolumn{1}{l|}{1.04e0}  & \multicolumn{1}{l|}{2.81e-2} & 2.36e-2  \\
\multicolumn{1}{l|}{Diffusion} & \multicolumn{1}{l|}{1.43e-4} & \multicolumn{1}{l|}{1.58e-1} & \multicolumn{1}{l|}{1.60e-1} & \multicolumn{1}{l|}{2.33e-3} & \multicolumn{1}{l|}{3.47e-4} & \multicolumn{1}{l|}{1.86e0}  & \multicolumn{1}{l|}{2.72e-2} & \multicolumn{1}{l|}{3.07e-4} & \colorbox{blue!25}{3.50e-5} \\
\multicolumn{1}{l|}{Heat}      & \multicolumn{1}{l|}{4.73e-3} & \multicolumn{1}{l|}{6.49e-3} & \multicolumn{1}{l|}{7.57e-3} & \multicolumn{1}{l|}{2.19e-3} & \multicolumn{1}{l|}{3.96e-3} & \multicolumn{1}{l|}{4.52e-1} & \multicolumn{1}{l|}{2.63e-1}  & \multicolumn{1}{l|}{1.16e-3} & \colorbox{blue!25}{4.10e-4} \\
\multicolumn{1}{l|}{Poisson}   & \multicolumn{1}{l|}{3.62e-3} & \multicolumn{1}{l|}{4.96e-1} & \multicolumn{1}{l|}{4.91e-1} & \multicolumn{1}{l|}{7.59e-4} & \multicolumn{1}{l|}{9.08e-4} & \multicolumn{1}{l|}{6.35e-1} & \multicolumn{1}{l|}{2.33e-1} & \multicolumn{1}{l|}{\colorbox{blue!25}{5.26e-4}} & 8.94e-4 \\
\multicolumn{1}{l|}{Burgers}   & \multicolumn{1}{l|}{1.86e-3} & \multicolumn{1}{l|}{5.59e-1} & \multicolumn{1}{l|}{5.36e-1} & \multicolumn{1}{l|}{7.49e-2} & \multicolumn{1}{l|}{1.30e-3} & \multicolumn{1}{l|}{9.94e-1} & \multicolumn{1}{l|}{7.52e-1} & \multicolumn{1}{l|}{2.95e-3} & \colorbox{blue!25}{3.16e-4} \\
\multicolumn{1}{l|}{NS} & \multicolumn{1}{l|}{5.26e-1} & \multicolumn{1}{l|}{7.14e-1} & \multicolumn{1}{l|}{6.33e-1} & \multicolumn{1}{l|}{6.94e-1} & \multicolumn{1}{l|}{3.77e-1} & \multicolumn{1}{l|}{5.46e-1}  & \multicolumn{1}{l|}{4.87e-1} & \multicolumn{1}{l|}{2.99e-1} & \colorbox{blue!25}{2.57e-1} \\
\midrule
\end{tabular}
\vskip -0.15in
\end{table*}

\begin{table}[ht]
\centering
\caption{$L2$ error of the Inverse Problems. * denotes added noises. Full results are in Appendix~\ref{Append: full results}.}
\label{tab: inverse}
\begin{tabular}{ccccc}
\toprule
                                                          & FF                             & SF                            & \textbf{RBF}                       & \textbf{RBF-P}  \\ \midrule
\multicolumn{1}{l|}{I-Burgers}  & \multicolumn{1}{l|}{2.39e-2} & \multicolumn{1}{l|}{2.43e-2} & \multicolumn{1}{l|}{1.74e-2} & \colorbox{blue!25}{1.57e-2} \\
\multicolumn{1}{l|}{I-Lorenz}   & \multicolumn{1}{l|}{6.51e-3}  & \multicolumn{1}{l|}{6.39e-3} & \multicolumn{1}{l|}{6.08e-3} & \colorbox{blue!25}{5.99e-3} \\ 
\multicolumn{1}{l|}{I-Burgers*}  & \multicolumn{1}{l|}{2.50e-2} & \multicolumn{1}{l|}{2.91e-2} & \multicolumn{1}{l|}{1.99e-2} & \colorbox{blue!25}{1.75e-2} \\
\multicolumn{1}{l|}{I-Lorenz*}   & \multicolumn{1}{l|}{7.93e-3}  & \multicolumn{1}{l|}{6.85e-3} & \multicolumn{1}{l|}{6.69e-3} & \colorbox{blue!25}{6.34e-3} \\ 
\midrule
\end{tabular}
\vskip -0.2in
\end{table}

\textbf{Time-dependent PDEs.}
Our solution in the Diffusion equation demonstrates superior performance compared to other methods by an order of magnitude. Boundary errors are notably more perceptible in Fourier-based methods, as illustrated in Appendix~\ref{Append: Qualitative}, Figure~\ref{fig:diffusion qualitative}. \\
The RBFs demonstrate enhanced capability in addressing multiscale problems, as illustrated in the Heat equation~\ref{equation: heat}. In the Heat equation formulation, there exists a substantial contrast in coefficients: $\frac{1}{{500\pi}^2}$ for the x-direction and $\frac{1}{{\pi}^2}$ for the y-direction. Figure~\ref{fig:heat qualitative} illustrates that the RBF method effectively preserves the details of the solution at each time step.\\
\textbf{Non-linear PDEs.} We assess the methods using two classic non-linear PDEs: the Burgers equation and the Navier-Stokes equation. Figure~\ref{fig:burgers qualitative} illustrates that RBFs with polynomial terms are better in addressing the discontinuity at $x=0$ in the Burgers equation. \\
\textbf{Inverse Problems.}
A major application of the PINNs is able to solve Inverse Problems. The unknown coefficients in the differential equations can be discovered by a small amount of data points. our methods have shown their efficacy in two Inverse Problems, shown in~\ref{tab: inverse} \\
Another experiment aimed to test the robustness of feature mapping functions to noise. $1\%$ Gaussian noises are added to the inverse Burgers problem and $0.5\%$ to the Lorenze system data. The results presented in Table~\ref{tab: inverse} reveal that the four tested feature mapping methods indicate a degree of immunity to noise. Furthermore, RBF-P stands out as the most resilient feature mapping function to noise. \\
All benchmarked method can be found in Appendix~\ref{Append: FM}.

\subsection{Ablation Study}

We carried out ablation studies on the number of RBFs in the feature mapping layer, the number of polynomials for RBF-P and different types of RBFs. Generally, a higher number of RBFs perform better but requires high computation resources. For different cases, the number of polynomials terms required varies. And among all test RBF functions, Gaussians present more stable results. The complete results can be found in Appendix~\ref{appendix: ablation study}. The complexity and scalability of different feature mapping functions are included in Appendix~\ref{appendix: complexity}.

\section{Conclusion and Future Work}
In conclusion, we have introduced a framework for designing an effective feature mapping function in PINNs and proposed Radial Basis Function-based approaches. Our method not only enhances generalisation across a range of forward and inverse physics problems but also surpasses other feature mapping methods by a substantial margin. The RBF feature mapping has the potential to be compatible with various other PINNs techniques, including novel activation functions and loss functions or training strategies such as curriculum training or causal training. While the primary focus of this work has been on solving Partial Differential Equations, the exploration of RBF feature mapping extends to its application in other coordinate-based input neural networks for different tasks.

\bibliography{main}

\begin{thebibliography}{36}
\providecommand{\natexlab}[1]{#1}
\providecommand{\url}[1]{\texttt{#1}}
\expandafter\ifx\csname urlstyle\endcsname\relax
  \providecommand{\doi}[1]{doi: #1}\else
  \providecommand{\doi}{doi: \begingroup \urlstyle{rm}\Url}\fi

\bibitem[Das \& Tesfamariam(2022)Das and Tesfamariam]{Das2022StateoftheArtRO}
Sourav Das and Solomon Tesfamariam.
\newblock State-of-the-art review of design of experiments for physics-informed deep learning.
\newblock \emph{ArXiv}, abs/2202.06416, 2022.

\bibitem[Daw et~al.(2022)Daw, Bu, Wang, Perdikaris, and Karpatne]{Daw22Propagation}
Arka Daw, Jie Bu, Sifan Wang, Paris Perdikaris, and Anuj Karpatne.
\newblock Mitigating propagation failures in physics-informed neural networks using retain-resample-release (r3) sampling.
\newblock In \emph{International Conference on Machine Learning}, 2022.

\bibitem[Farazandeh \& Mirzaei(2021)Farazandeh and Mirzaei]{Farazandeh2021ARR}
Elham Farazandeh and Davoud Mirzaei.
\newblock A rational rbf interpolation with conditionally positive definite kernels.
\newblock \emph{Advances in Computational Mathematics}, 47, 2021.

\bibitem[Giladi et~al.(2021)Giladi, Ben-Haim, Nevo, Matias, and Soudry]{Giladi2021PhysicsAwareDW}
Niv Giladi, Zvika Ben-Haim, Sella Nevo, Yossi Matias, and Daniel Soudry.
\newblock Physics-aware downsampling with deep learning for scalable flood modeling.
\newblock In \emph{Neural Information Processing Systems}, 2021.

\bibitem[Goyeneche et~al.(2023)Goyeneche, Ramachandran, Wang, Karasan, Cheng, Stella, and Lustig]{Goyeneche2023ResoNet}
Alfredo~De Goyeneche, Shreya Ramachandran, Ke~Wang, Ekin Karasan, Joseph~Yitan Cheng, X~Yu Stella, and Michael Lustig.
\newblock Resonet: Noise-trained physics-informed mri off-resonance correction.
\newblock In \emph{Neural Information Processing Systems}, 2023.

\bibitem[Hornik et~al.(1989)Hornik, Stinchcombe, and White]{Hornik1989MultilayerFN}
Kurt Hornik, Maxwell~B. Stinchcombe, and Halbert~L. White.
\newblock Multilayer feedforward networks are universal approximators.
\newblock \emph{Neural Networks}, 2:\penalty0 359--366, 1989.

\bibitem[Hu et~al.(2023)Hu, Li, Klemme, Nam, Ma, Amrouch, and Xiong]{Hu2023SyncTREE}
Yuting Hu, Jiajie Li, Florian Klemme, Gi-Joon Nam, Tengfei Ma, Hussam Amrouch, and Jinjun Xiong.
\newblock Synctree: Fast timing analysis for integrated circuit design through a physics-informed tree-based graph neural network.
\newblock In \emph{Neural Information Processing Systems}, 2023.

\bibitem[Jagtap \& Karniadakis(2019)Jagtap and Karniadakis]{Jagtap2019AdaptiveAF}
Ameya~Dilip Jagtap and George~Em Karniadakis.
\newblock Adaptive activation functions accelerate convergence in deep and physics-informed neural networks.
\newblock \emph{J. Comput. Phys.}, 404, 2019.

\bibitem[Jagtap et~al.(2020)Jagtap, Kawaguchi, and Karniadakis]{Jagtap2020LocallyAA}
Ameya~Dilip Jagtap, Kenji Kawaguchi, and George~Em Karniadakis.
\newblock Locally adaptive activation functions with slope recovery for deep and physics-informed neural networks.
\newblock \emph{Proceedings of the Royal Society A}, 476, 2020.

\bibitem[Karniadakis et~al.(2021)Karniadakis, Kevrekidis, Lu, Perdikaris, Wang, and Yang]{Karniadakis2021PhysicsinformedML}
George~Em Karniadakis, Ioannis~G. Kevrekidis, Lu~Lu, Paris Perdikaris, Sifan Wang, and Liu Yang.
\newblock Physics-informed machine learning.
\newblock \emph{Nature Reviews Physics}, 3:\penalty0 422 -- 440, 2021.

\bibitem[Kashinath et~al.(2021)Kashinath, Mustafa, Albert, Wu, Jiang, Esmaeilzadeh, Azizzadenesheli, Wang, Chattopadhyay, Singh, et~al.]{kashinath2021physics}
Karthik Kashinath, M~Mustafa, Adrian Albert, JL~Wu, C~Jiang, Soheil Esmaeilzadeh, Kamyar Azizzadenesheli, R~Wang, A~Chattopadhyay, A~Singh, et~al.
\newblock Physics-informed machine learning: case studies for weather and climate modelling.
\newblock \emph{Philosophical Transactions of the Royal Society A}, 379\penalty0 (2194):\penalty0 20200093, 2021.

\bibitem[Mildenhall et~al.(2020)Mildenhall, Srinivasan, Tancik, Barron, Ramamoorthi, and Ng]{Mildenhall2020NeRFRS}
Ben Mildenhall, Pratul~P. Srinivasan, Matthew Tancik, Jonathan~T. Barron, Ravi Ramamoorthi, and Ren Ng.
\newblock Nerf: Representing scenes as neural radiance fields for view synthesis.
\newblock \emph{Commun. ACM}, 65:\penalty0 99--106, 2020.

\bibitem[Nabian et~al.(2021)Nabian, Gladstone, and Meidani]{Nabian2021EfficientTO}
Mohammad~Amin Nabian, Rini~Jasmine Gladstone, and Hadi Meidani.
\newblock Efficient training of physics‐informed neural networks via importance sampling.
\newblock \emph{Computer‐Aided Civil and Infrastructure Engineering}, 36:\penalty0 962 -- 977, 2021.

\bibitem[Ni \& Qureshi(2023)Ni and Qureshi]{Ni2022NTFieldsNT}
Ruiqi Ni and Ahmed~Hussain Qureshi.
\newblock Ntfields: Neural time fields for physics-informed robot motion planning.
\newblock In \emph{International Conference on Learning Representations}, 2023.

\bibitem[Nicoli et~al.(2023)Nicoli, Anders, Funcke, Hartung, Jansen, Kuhn, Muller, Stornati, Kessel, and Nakajima]{Nicoli2023Quant}
Kim~Andrea Nicoli, Christopher~J Anders, Lena Funcke, Tobias Hartung, Karl Jansen, Stefan Kuhn, Klaus~Robert Muller, Paolo Stornati, Pan Kessel, and Shinichi Nakajima.
\newblock Physics-informed bayesian optimization of variational quantum circuits.
\newblock In \emph{Neural Information Processing Systems}, 2023.

\bibitem[Pokkunuru et~al.(2023)Pokkunuru, Rooshenas, Strauss, Abhishek, and Khan]{pokkunuru2023improved}
Akarsh Pokkunuru, Amirmohmmad Rooshenas, Thilo Strauss, Anuj Abhishek, and Taufiquar Khan.
\newblock Improved training of physics-informed neural networks using energy-based priors: a study on electrical impedance tomography.
\newblock In \emph{International Conference on Learning Representations}, 2023.

\bibitem[Rahaman et~al.(2018)Rahaman, Baratin, Arpit, Dr{\"a}xler, Lin, Hamprecht, Bengio, and Courville]{Rahaman2018OnTS}
Nasim Rahaman, Aristide Baratin, Devansh Arpit, Felix Dr{\"a}xler, Min Lin, Fred~A. Hamprecht, Yoshua Bengio, and Aaron~C. Courville.
\newblock On the spectral bias of neural networks.
\newblock In \emph{International Conference on Machine Learning}, 2018.

\bibitem[Rahimi \& Recht(2007)Rahimi and Recht]{Rahimi2007RandomFF}
Ali Rahimi and Benjamin Recht.
\newblock Random features for large-scale kernel machines.
\newblock In \emph{Neural Information Processing Systems}, 2007.

\bibitem[Raissi et~al.(2019)Raissi, Perdikaris, and Karniadakis]{Raissi2019PhysicsinformedNN}
Maziar Raissi, Paris Perdikaris, and George~Em Karniadakis.
\newblock Physics-informed neural networks: A deep learning framework for solving forward and inverse problems involving nonlinear partial differential equations.
\newblock \emph{J. Comput. Phys.}, 378:\penalty0 686--707, 2019.

\bibitem[Ramasinghe \& Lucey(2021)Ramasinghe and Lucey]{Ramasinghe2021ALR}
Sameera Ramasinghe and Simon Lucey.
\newblock A learnable radial basis positional embedding for coordinate-mlps.
\newblock In \emph{AAAI Conference on Artificial Intelligence}, 2021.

\bibitem[Ramasinghe \& Lucey(2022)Ramasinghe and Lucey]{Ramasinghe2022Periodicity}
Sameera Ramasinghe and Simon Lucey.
\newblock Beyond periodicity: Towards a unifying framework for activations in coordinate-mlps.
\newblock In \emph{ECCV 2022: 17th European Conference Proceedings, Part XXXIII}, Berlin, Heidelberg, 2022. Springer-Verlag.
\newblock ISBN 978-3-031-19826-7.

\bibitem[Salehi \& Giannacopoulos(2022)Salehi and Giannacopoulos]{Salehi2021PhysGNNAP}
Yasmin Salehi and Dennis~D. Giannacopoulos.
\newblock Physgnn: A physics-driven graph neural network based model for predicting soft tissue deformation in image-guided neurosurgery.
\newblock In \emph{Neural Information Processing Systems}, 2022.

\bibitem[Saragadam et~al.(2023)Saragadam, LeJeune, Tan, Balakrishnan, Veeraraghavan, and Baraniuk]{Saragadam2023WIREWI}
Vishwanath Saragadam, Daniel LeJeune, Jasper Tan, Guha Balakrishnan, Ashok Veeraraghavan, and Richard Baraniuk.
\newblock Wire: Wavelet implicit neural representations.
\newblock \emph{2023 IEEE/CVF Conference on Computer Vision and Pattern Recognition (CVPR)}, pp.\  18507--18516, 2023.

\bibitem[Sitzmann et~al.(2020)Sitzmann, Martel, Bergman, Lindell, and Wetzstein]{Sitzmann2020ImplicitNR}
Vincent Sitzmann, Julien N.~P. Martel, Alexander~W. Bergman, David~B. Lindell, and Gordon Wetzstein.
\newblock Implicit neural representations with periodic activation functions.
\newblock In \emph{Neural Information Processing Systems}, 2020.

\bibitem[Smith et~al.(2022)Smith, Seccamonte, Swami, and Bullo]{Smith2022PhysicsInformedIR}
Kevin~D. Smith, Francesco Seccamonte, Ananthram Swami, and Francesco Bullo.
\newblock Physics-informed implicit representations of equilibrium network flows.
\newblock In \emph{Neural Information Processing Systems}, 2022.

\bibitem[Tancik et~al.(2020)Tancik, Srinivasan, Mildenhall, Fridovich-Keil, Raghavan, Singhal, Ramamoorthi, Barron, and Ng]{Tancik2020FourierFL}
Matthew Tancik, Pratul~P. Srinivasan, Ben Mildenhall, Sara Fridovich-Keil, Nithin Raghavan, Utkarsh Singhal, Ravi Ramamoorthi, Jonathan~T. Barron, and Ren Ng.
\newblock Fourier features let networks learn high frequency functions in low dimensional domains.
\newblock In \emph{Neural Information Processing Systems}, 2020.

\bibitem[Thangamuthu et~al.(2022)Thangamuthu, Kumar, Bishnoi, Bhattoo, Krishnan, and Ranu]{Thangamuthu2022Dynamical}
Abishek Thangamuthu, Gunjan Kumar, Suresh Bishnoi, Ravinder Bhattoo, N~M~Anoop Krishnan, and Sayan Ranu.
\newblock Unravelling the performance of physics-informed graph neural networks for dynamical systems.
\newblock In \emph{Neural Information Processing Systems}, 2022.

\bibitem[Wang et~al.(2021{\natexlab{a}})Wang, Liu, Yang, and Tong]{Wang2021SplinePE}
Peng-Shuai Wang, Yang Liu, Yu-Qi Yang, and Xin Tong.
\newblock Spline positional encoding for learning 3d implicit signed distance fields.
\newblock In \emph{International Joint Conference on Artificial Intelligence}, 2021{\natexlab{a}}.

\bibitem[Wang et~al.(2020)Wang, Yu, and Perdikaris]{Wang2020WhenAW}
Sifan Wang, Xinling Yu, and Paris Perdikaris.
\newblock When and why pinns fail to train: A neural tangent kernel perspective.
\newblock \emph{J. Comput. Phys.}, 449:\penalty0 110768, 2020.

\bibitem[Wang et~al.(2021{\natexlab{b}})Wang, Wang, and Perdikaris]{WANG2021113938}
Sifan Wang, Hanwen Wang, and Paris Perdikaris.
\newblock On the eigenvector bias of fourier feature networks: From regression to solving multi-scale pdes with physics-informed neural networks.
\newblock \emph{Computer Methods in Applied Mechanics and Engineering}, 384:\penalty0 113938, 2021{\natexlab{b}}.
\newblock ISSN 0045-7825.

\bibitem[Wang et~al.(2023)Wang, Sankaran, Wang, and Perdikaris]{Wang2023AnEG}
Sifan Wang, Shyam Sankaran, Hanwen Wang, and Paris Perdikaris.
\newblock An expert's guide to training physics-informed neural networks.
\newblock \emph{ArXiv}, abs/2308.08468, 2023.

\bibitem[Wong et~al.(2022)Wong, Ooi, Gupta, and Ong]{Wong2022Sinusoidal}
Jian Wong, Chinchun Ooi, Abhishek Gupta, and Yew Ong.
\newblock Learning in sinusoidal spaces with physics-informed neural networks.
\newblock \emph{IEEE Transactions on Artificial Intelligence}, PP:\penalty0 1--5, 01 2022.
\newblock \doi{10.1109/TAI.2022.3192362}.

\bibitem[Wu et~al.(2023)Wu, Zhu, Tan, Kartha, and Lu]{WU2023115671}
Chenxi Wu, Min Zhu, Qinyang Tan, Yadhu Kartha, and Lu~Lu.
\newblock A comprehensive study of non-adaptive and residual-based adaptive sampling for physics-informed neural networks.
\newblock \emph{Computer Methods in Applied Mechanics and Engineering}, 403:\penalty0 115671, 2023.
\newblock ISSN 0045-7825.

\bibitem[Xie et~al.(2021)Xie, Takikawa, Saito, Litany, Yan, Khan, Tombari, Tompkin, Sitzmann, and Sridhar]{Xie2021NeuralFI}
Yiheng Xie, Towaki Takikawa, Shunsuke Saito, Or~Litany, Shiqin Yan, Numair Khan, Federico Tombari, James Tompkin, Vincent Sitzmann, and Srinath Sridhar.
\newblock Neural fields in visual computing and beyond.
\newblock \emph{Computer Graphics Forum}, 41, 2021.

\bibitem[Yang et~al.(2023)Yang, Qiu, and Fu]{Yang23DMIS}
Zijiang Yang, Zhongwei Qiu, and Dongmei Fu.
\newblock Dmis: Dynamic mesh-based importance sampling for training physics-informed neural networks.
\newblock AAAI Press, 2023.

\bibitem[Zheng et~al.(2022)Zheng, Ramasinghe, Li, and Lucey]{Zheng2021RethinkingPE}
Jianqiao Zheng, Sameera Ramasinghe, Xueqian Li, and Simon Lucey.
\newblock Trading positional complexity vs. deepness in coordinate networks.
\newblock \emph{Proceedings of the European Conference on Computer Vision (ECCV)}, 2022.

\end{thebibliography}
\bibliographystyle{iclr2024_workshop}

\newpage
\appendix
\section{Related Work}
\textbf{Coordinate Sampling.}
As a mesh-free method, PINNs are normally evaluated on scattered collocation points both on the interior domain and IC/BC.  Therefore, the sampling strategy is crucial to PINNs' performance and efficiency. A poorly distributed initial sampling can lead to the PDE system being ill-conditioned and NN training instability. The whole design of experiments on the fixed input sampling is reviewed by~\cite{Das2022StateoftheArtRO}. Based on the study of uniform sampling, ~\citet{WU2023115671} proposed an adaptive sampling scheme that refines high residual area during training. Similarly, Importance Sampling inspired by Monte Carlo approximation is investigated by~\cite{Nabian2021EfficientTO, Yang23DMIS}.~\citet{Daw22Propagation} proposed a novel sampling strategy that mitigates the `propagation failure' of solutions from IC/BC to the PDE residual field.\\
\textbf{Novel Activation.}
The activation function in the MLP has been found to play an important role in the convergence of the PINNs. Popular activation ReLU is deficient for high-order PDEs since its second-order derivative is 0. Apart from the standard Tanh activation~\cite{Raissi2019PhysicsinformedNN}, layer-wise and neuron-wise adaptive activation are proven to be useful to accelerate the training~\cite{Jagtap2019AdaptiveAF, Jagtap2020LocallyAA}. Another line of seminal work, SIREN~\cite{Sitzmann2020ImplicitNR}, which uses periodic activation function, has achieved remarkable results in  Neural Representation and tested on solving the Poisson equation. Gaussian~\cite{Ramasinghe2022Periodicity} and Gabor Wavelet activations~\cite{, Saragadam2023WIREWI} are proven to be effective alternatives.\\
\textbf{Positional Embedding.} Broadly speaking, PINNs can also be considered as a special type of Neural Fields~\cite{Xie2021NeuralFI} in visual computing, which specifically feed coordinate-based input to MLPs that represent continuous field quantity (e.g. velocity field in fluid mechanics) over arbitrary spatial and temporal resolution. However, the PINNs community often ignores the fact both perspectives function the same way as Implicit Neural Representations. In the Neural Field, images and 3D shapes are naturally high-frequency signals, whereas deep networks are inherently learning towards the low-frequency components~\cite{Rahaman2018OnTS}. Feature mapping hence has become a standard process in practice that maps the low-dimension coordinates to high-dimension space. The pioneering work was conducted by~\cite{Rahimi2007RandomFF}, who used Fourier features to approximate any stationary kernel principled by Bochner's theorem. the derivative works are done such as Positional Encoding~\cite{Mildenhall2020NeRFRS}, Random Feature~\cite{Tancik2020FourierFL} and Sinusoidal Feature~\cite{Sitzmann2020ImplicitNR}. Another concurrent work discusses non-periodic feature mapping~\cite{Zheng2021RethinkingPE, Ramasinghe2021ALR, Wang2021SplinePE}. To the best of our knowledge, feature mapping in PINNs has been largely uninvestigated. Only a few work preliminarily adopted Fourier-feature-based methods in PINN~\cite{WANG2021113938, Wang2023AnEG, Wong2022Sinusoidal}. 

\section{Benchmarked feature mapping methods}
\label{Append: FM}
\textbf{Basic Encoding:}~\cite{Mildenhall2020NeRFRS} $\varphi(x) = [cos(2\pi\sigma^{j/m}x, sin(2\pi\sigma^{j/m}x]^T$ for $j = 0, .., m-1$.\\
\textbf{Positional Encoding:}~\cite{Mildenhall2020NeRFRS} $\varphi(x) = [cos(2\pi\sigma^{j/m}x, sin(2\pi\sigma^{j/m}x]^T$ for $j = 0, .., m-1$.\\
\textbf{Random Fourier:}~\cite{Tancik2020FourierFL} $\varphi(x) = [cos(2\pi \sigma \mathcal{B} x), sin(2\pi \sigma \mathcal{B} x)]^T$, where $\mathcal{B} \in \mathbb{R}^{m\times d}$ is sampled from $\mathcal{N}(0, 1)$ and $\sigma$ is an arbitrary scaling factor varies case to case.\\
\textbf{Sinusoidal  Feature:}~\cite{Sitzmann2020ImplicitNR} $\varphi(x) = [sin(2\pi \mathbf{W} x + \mathbf{b})]^T$, where $\mathbf{W}$ and $\mathbf{b}$ are trainable parameters.\\
\textbf{Complex Triangle:}~\cite{Zheng2021RethinkingPE} $\varphi(x) = [max(1-\frac{|x_1-t|}{0.5d}, 0), max(1-\frac{|x_2-t|}{0.5d}, 0), \cdots, max(1-\frac{|x_i-t|}{0.5d}, 0)]^T$, where $t$ is uniformly sampled from 0 to 1.\\
\textbf{Complex Gaussian:}~\cite{Zheng2021RethinkingPE} $\varphi(x) = [e^{-0.5(x_1 - \tau/d)^2/\sigma^2}\bigotimes\cdots\bigotimes e^{-0.5(x_d - \tau/d)^2/\sigma^2}]^T$, where $\tau$ is uniformly sampled from $[0, 1]$, and $\bigotimes$ is the Kronecker product. \\

\section{Spectral Bias and Composed NTK}
\label{Append: Spectral}
\subsection{Spectral Bias}
Normally PINNs are setup as a standard MLP model $f(\mathbf{x};\theta)$, and $\theta$ is optimized on the loss function $L(\theta) = \frac{1}{2}\left |f(\mathbf{x}; \theta) - Y\right|^2= \frac{1}{2}\sum_{i}^{N}(f(x_i; \theta)-y_i)^2$, where $X$, $Y$ and $\theta$ are training input, training ground truth and model parameters. For an easier formulation, we replace the conventional gradient descent formulation $\theta_{t+1} = \theta_t - \alpha \nabla_{\theta}L(\theta_t)$ to a gradient flow equation:
\begin{equation}
\frac{d\theta}{dt} = - \alpha \nabla_{\theta}L(\theta_t)
\end{equation}
where $\alpha$ should be an infinitesimally small learning rate in the NTK setting. \\
Given PDE collocation data points$\{x_r^i, \mathcal{D}(\hat u_\theta(x_{r}^i)) \}_{i = 1}^{N_r}$, and boundary training points$\{x_{bc}^i, \mathcal{B}(\hat u_\theta(x_{bc}^i))\}_{i = 1}^{N_{bc}}$. The gradient flow can be formulated as~\cite{Wang2020WhenAW}:
\begin{equation}
\label{equa: pinn gradient flow}
\left[\begin{array}{c}
\frac{d u\left(x_b, \theta_t\right)}{d t} \\
\frac{d \mathcal{L} u\left(x_r, \theta_t\right)}{d t}
\end{array}\right] = -\left[\begin{array}{ll}
\boldsymbol{K}_{u u}^t & \boldsymbol{K}_{u r}^t \\
\boldsymbol{K}_{r u}^t & \boldsymbol{K}_{r r}^t
\end{array}\right] \cdot \left[\begin{array}{c}
u\left(x_b, \theta_t\right)- \mathcal{B}(\hat u_\theta(x_{bc})) \\
\mathcal{L} u\left(x_r, \theta_t\right)-\mathcal{D}(\hat u_\theta(x_{r}))
\end{array}\right],
\end{equation}
where the Kernels $\boldsymbol{K}$ are:
\begin{equation}
\begin{aligned}
& \left(\boldsymbol{K}_{u u}^t\right)_{i j}=\left\langle\frac{d u\left(x_b^i, \theta_t\right)}{d \theta}, \frac{d u\left(x_b^j, \theta_t\right)}{d \theta}\right\rangle \\
& \left(\boldsymbol{K}_{r r}^t\right)_{i j}=\left\langle\frac{d \mathcal{L}\left(x_r^i, \theta_t\right)}{d \theta}, \frac{d \mathcal{L}\left(x_r^j, \theta_t\right)}{d \theta}\right\rangle\\
& \left(\boldsymbol{K}_{u r}^t\right)_{i j}=\left(\boldsymbol{K}_{r u}^t\right)_{i j}=\left\langle\frac{d u\left(x_b^i, \theta_t\right)}{d \theta}, \frac{d \mathcal{L} u\left(x_r^j, \theta_t\right)}{d \theta}\right\rangle 
\end{aligned}
\end{equation}
Since $\boldsymbol{K}$ remains stationary, then $\boldsymbol{K}^t \approx \boldsymbol{K}^0$ as NN width tends to infinity, Equation~\ref{equa: pinn gradient flow} is rewritten as:
\begin{equation}
\begin{aligned}
\left[\begin{array}{c}
\frac{d u\left(x_b, \theta_t\right)}{d t} \\
\frac{d \mathcal{L} u\left(x_r, \theta_t\right)}{d t}
\end{array}\right] & \approx - \boldsymbol{K}^0 \left[\begin{array}{c}
u\left(x_b, \theta_t\right)- \mathcal{B}(\hat u_\theta(x_{bc})) \\
\mathcal{L} u\left(x_r, \theta_t\right)-\mathcal{D}(\hat u_\theta(x_{r}))
\end{array}\right] \\
& \approx(I - e^{-\boldsymbol{K}^0t}) \cdot\left[\begin{array}{l}
\mathcal{B}(\hat u_\theta(x_{bc}) \\
\mathcal{D}(\hat u_\theta(x_{r}))
\end{array}\right]
\end{aligned}
\end{equation}
By Schur product theorem, $\boldsymbol{K}^0$ is always Positive Semi-definite, hence it can be Eigen-decomposed to $\boldsymbol{Q}^T \Lambda \boldsymbol{Q}$, where $\boldsymbol{Q}$ is an orthogonal matrix and $\Lambda$ is a diagonal matrix with eigenvalues $\lambda _i$ in the entries. We can rearrange the training error in the form of:
\begin{equation}
\begin{aligned}
\left[\begin{array}{c}
\frac{d u\left(x_b, \theta_t\right)}{d t} \\
\frac{d \mathcal{L} u\left(x_r, \theta_t\right)}{d t}
\end{array}\right] - \left[\begin{array}{l}
\mathcal{B}(\hat u_\theta(x_{bc}) \\
\mathcal{D}(\hat u_\theta(x_{r}))
\end{array}\right] & \approx (I - e^{-\boldsymbol{K}^0t}) \cdot \left[\begin{array}{l}
\mathcal{B}(\hat u_\theta(x_{bc}) \\
\mathcal{D}(\hat u_\theta(x_{r}))
\end{array}\right] - \left[\begin{array}{l}
\mathcal{B}(\hat u_\theta(x_{bc}) \\
\mathcal{D}(\hat u_\theta(x_{r}))
\end{array}\right] \\
& \approx -\boldsymbol{Q}^T e^{-\Lambda t} \boldsymbol{Q} \cdot \left[\begin{array}{l}
\mathcal{B}(\hat u_\theta(x_{bc}) \\
\mathcal{D}(\hat u_\theta(x_{r}))
\end{array}\right] \\
\end{aligned}
\end{equation}
where $e^{-\Lambda t} = \left[\begin{array}{cccc} e^{-\lambda_1 t} & & \\
& \ddots & \\
& & e^{-\lambda_N t}
\end{array}\right] $. This indicates the decrease of training error in each component is exponentially proportional to the eigenvalues of the deterministic NTK, and the NN is inherently biased to learn along larger eigenvalues entries of the $\boldsymbol{K}^0$.
\newpage

\subsection{Composed NTK}
The Fourier feature layer is defined as:\\
\begin{equation}
\varphi(\mathbf{x})=\left[a_1 \cos \left(2 \pi b_1^{\mathrm{T}} \mathbf{x}\right), a_1 \sin \left(2 \pi \mathbf{b}_1^{\mathrm{T}} \mathbf{x}\right), \ldots, a_m \cos \left(2 \pi \mathbf{b}_m^{\mathrm{T}} \mathbf{x}\right), a_m \sin \left(2 \pi \mathbf{b}_m^{\mathrm{T}} \mathbf{x}\right)\right]^{\mathrm{T}}
\end{equation}\\
Hence the NTK is computed by:\\
\begin{equation}
\begin{aligned}
\boldsymbol{K}_{\Phi}\left(x_i, x_j\right)  & = \varphi(x_i)^T\varphi(x_j) \\
& =\left[\begin{array}{c}
A_k\cos \left(2\pi \boldsymbol{b_m} x_i\right) \\
A_k\sin \left(2\pi \boldsymbol{b_m} x_j\right)
\end{array}\right]^{\mathrm{T}} \cdot\left[\begin{array}{c}
A_k\cos \left(2\pi \boldsymbol{b_m} x_i\right) \\
A_k\sin \left(2\pi \boldsymbol{b_m} x_j\right)
\end{array}\right] \\
& =  \sum_{k=1}^m A_k \cos \left(2\pi b_k^T x_i\right) \cos \left(2\pi b_k^T x_j\right) \\
&+A_k \sin \left(2\pi b_k^T x_i\right) \sin \left(2\pi b_k^T x_j\right) \\
&\boxed{\begin{aligned}
                           \text{Trigonometric Identities:} \cos(c - d)&= \cos c\cos d+\sin c \sin d
                      \end{aligned}} \\
& =  \sum_{k=1}^m A_k^2 \cos \left(2\pi b_k^T\left(x_i-x_j\right)\right) .
\end{aligned}
\end{equation}\\
where $A$ is the Fourier Series coefficients, $ \boldsymbol{b} $ is randomly sampled from $\mathcal{N}(0,\,\sigma^{2})$ and $\sigma$ is an arbitrary hyperparameter that controls the bandwidth.
Thereafter, the feature space becomes the input of the NTK which gives the identities: $\boldsymbol{K}_{NTK}(x_i^Tx_j) = \boldsymbol{K}_{NTK}(\varphi(x_i)^T\varphi(x_j)) = \boldsymbol{K}_{NTK}(\boldsymbol{K}_{\Phi}(x_i - x_j))$.

\newpage
\section{Complete experimental results for Table~\ref{tab: forward pdes}\&\ref{tab: inverse}}
\label{Append: full results}
\subsection{Complete results for Table~\ref{tab: forward pdes}}
\begin{table}[H]
\caption{Full PDEs benchmark results comparing different feature mapping methods in $L2$ error. The best results are in \colorbox{blue!25}{Blue}. Standard deviations are shown after $\pm$.}
\label{tab: full forward pdes}
\centering
\small
\begin{tabular}{@{}cccccc@{}}
\toprule
                               & PINN                           & BE                 & PE            & FF                & SF               \\ \midrule
\multicolumn{1}{l|}{Wave}      & \multicolumn{1}{l|}{3.73e-1$\pm$2.37e-2} & \multicolumn{1}{l|}{1.04e0$\pm$3.55e-1}  & \multicolumn{1}{l|}{1.01e0$\pm$4.02e-1}  & \multicolumn{1}{l|}{\colorbox{blue!25}{2.38e-3$\pm$3.75e-4}} & \multicolumn{1}{l|}{7.93e-3$\pm$9.32e-4}  \\
\multicolumn{1}{l|}{Diffusion} & \multicolumn{1}{l|}{1.43e-4$\pm$4.84e-5} & \multicolumn{1}{l|}{1.58e-1$\pm$6.13e-2} & \multicolumn{1}{l|}{1.60e-1$\pm$1.20e-2} & \multicolumn{1}{l|}{2.33e-3$\pm$7.51e-4} & \multicolumn{1}{l|}{3.47e-4$\pm$6.11e-5} \\
\multicolumn{1}{l|}{Heat}      & \multicolumn{1}{l|}{4.73e-3$\pm$6.14e-5} & \multicolumn{1}{l|}{6.49e-3$\pm$6.37e-4} & \multicolumn{1}{l|}{7.57e-3$\pm$1.02e-4} & \multicolumn{1}{l|}{2.19e-3$\pm$3.12e-4} & \multicolumn{1}{l|}{3.96e-3$\pm$2.56e-4} \\
\multicolumn{1}{l|}{Poisson}   & \multicolumn{1}{l|}{3.62e-3$\pm$1.24e-4} & \multicolumn{1}{l|}{4.96e-1$\pm$2.15e-2} & \multicolumn{1}{l|}{4.91e-1$\pm$1.08e-2} & \multicolumn{1}{l|}{7.58e-4$\pm$9.01e-5}  & \multicolumn{1}{l|}{9.07e-4$\pm$1.02e-5} \\
\multicolumn{1}{l|}{Burgers}   & \multicolumn{1}{l|}{1.86e-3$\pm$1.20e-4} & \multicolumn{1}{l|}{5.58e-1$\pm$2.57e-2} & \multicolumn{1}{l|}{5.36e-1$\pm$3.70e-2} & \multicolumn{1}{l|}{7.50e-2$\pm$5.15e-3} & \multicolumn{1}{l|}{1.30e-3$\pm$6.21e-4}   \\
\multicolumn{1}{l|}{Steady NS} & \multicolumn{1}{l|}{5.26e-1$\pm$1.01e-2} & \multicolumn{1}{l|}{7.14e-1$\pm$1.33e-2} & \multicolumn{1}{l|}{6.33e-1$\pm$2.35e-2} & \multicolumn{1}{l|}{6.94e-1$\pm$1.06e-3} & \multicolumn{1}{l|}{3.77e-1$\pm$2.37e-2} \\
\midrule
\end{tabular}
\centering
\small
\begin{tabular}{@{}ccccc@{}}
\toprule
          & CT                     & CG                       & \textbf{RBF}                            & \textbf{RBF-P}   \\ \midrule
\multicolumn{1}{l|}{Wave}     & \multicolumn{1}{l|}{1.11e0$\pm$3.21e-2}  & \multicolumn{1}{l|}{1.03e0$\pm$1.05e-2}  & \multicolumn{1}{l|}{2.81e-2$\pm$3.67e-3} & 2.36e-2$\pm$1.59e-2  \\
\multicolumn{1}{l|}{Diffusion}& \multicolumn{1}{l|}{1.86e0$\pm$2.31e-2}  & \multicolumn{1}{l|}{2.72e-2$\pm$1.02e-1} & \multicolumn{1}{l|}{3.06e-4$\pm$9.51e-6} & \colorbox{blue!25}{3.49e-5$\pm$6.54e-6} \\
\multicolumn{1}{l|}{Heat}     & \multicolumn{1}{l|}{4.52e-1$\pm$6.51e-2} & \multicolumn{1}{l|}{2.62e-1$\pm$2.36e-2}  & \multicolumn{1}{l|}{1.15e-3$\pm$1.02e-4} & \colorbox{blue!25}{4.09e-4$\pm$9.62e-6} \\
\multicolumn{1}{l|}{Poisson}   & \multicolumn{1}{l|}{6.34e-1$\pm$3.04e-1} & \multicolumn{1}{l|}{2.33e-1$\pm$5.47e-2} & \multicolumn{1}{l|}{\colorbox{blue!25}{5.25e-4$\pm$6.24e-5}} & 8.94e-4$\pm$ 6.51e-5\\
\multicolumn{1}{l|}{Burgers}   & \multicolumn{1}{l|}{9.93e-1$\pm$4.51e-2} & \multicolumn{1}{l|}{7.52e-1$\pm$3.24e-2} & \multicolumn{1}{l|}{2.94e-3$\pm$2.35e-4} & \colorbox{blue!25}{3.15e-4$\pm$2.14e-5} \\
\multicolumn{1}{l|}{Steady NS}  & \multicolumn{1}{l|}{5.46e-1$\pm$2.35e-2}  & \multicolumn{1}{l|}{4.86e-1$\pm$3.65e-2} & \multicolumn{1}{l|}{2.99e-1$\pm$6.51e-2} & \colorbox{blue!25}{2.56e-1$\pm$6.21e-2} \\
\midrule
\end{tabular}
\end{table}

\subsection{Complete results for Table~\ref{tab: inverse}}
\begin{table}[H]
\caption{Full Benchmark results on the Inverse problems in $L2$ error. * indicates problems with noises added to the data.}
\label{tab: full inverse}
\centering
\small
\begin{tabular}{lllll}
\toprule
                                                          & FF                             & SF                            & \textbf{RBF}                       & \textbf{RBF-P}  \\ \midrule
\multicolumn{1}{l|}{I-Burgers}  & \multicolumn{1}{l|}{2.39e-2$\pm$9.64e-4} & \multicolumn{1}{l|}{2.43e-2$\pm$4.678e-3} & \multicolumn{1}{l|}{1.74e-2$\pm$6.57e-3} & \colorbox{blue!25}{1.57e-2$\pm$9.36e-4} \\
\multicolumn{1}{l|}{I-Lorenz}   & \multicolumn{1}{l|}{6.51e-3$\pm$7.65e-4}  & \multicolumn{1}{l|}{6.39e-3$\pm$6.21e-4} & \multicolumn{1}{l|}{6.08e-3$\pm$3.69e-4} & \colorbox{blue!25}{5.99e-3$\pm$2.31e-4} \\ 
\multicolumn{1}{l|}{I-Burgers*}  & \multicolumn{1}{l|}{2.50e-2$\pm$6.32e-3} & \multicolumn{1}{l|}{2.91e-2$\pm$2.69e-3} & \multicolumn{1}{l|}{1.99e-2$\pm$3.62e-3} & \colorbox{blue!25}{1.75e-2$\pm$5.63e-3} \\
\multicolumn{1}{l|}{I-Lorenz*}   & \multicolumn{1}{l|}{7.93e-3$\pm$8.65e-4}  & \multicolumn{1}{l|}{6.85e-3$\pm$6.36e-4} & \multicolumn{1}{l|}{6.69e-3$\pm$5.20e-4} & \colorbox{blue!25}{6.34e-3$\pm$8.61e-4} \\ 
\midrule
\end{tabular}
\end{table}

\newpage

\section{Ablation Study}
\label{appendix: ablation study}
In this section, we show some additional experiments on our RBF feature mapping including investigations on the Number of RBFs, Number of Polynomials and different RBF types.
\subsection{Number of RBFs}
\begin{figure}[ht]
\begin{center}
\centerline{\includegraphics[height=5cm]{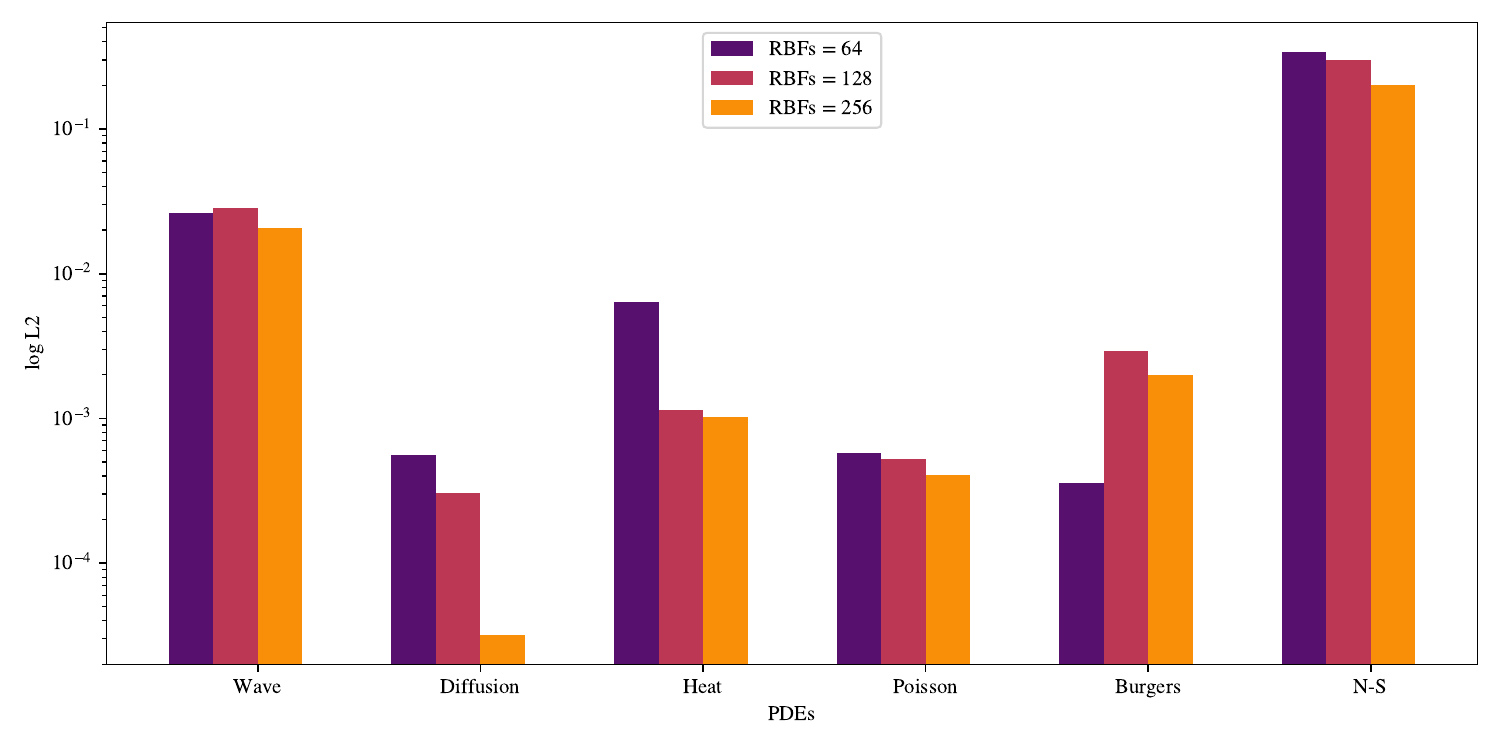}}
\caption{Ablation study on different number of RBFs}
\label{ablation: rbf number}
\end{center}
\end{figure}

Figure~\ref{ablation: rbf number} has shown generally more RBFs (256) yield better results. It however does demand a higher memory and can be slow in some cases.  It shows in the Diffusion equation, with 256 RBFs, the error reduces quite significantly. Otherwise, it only has limited improvements because the error is already very low. We use 128 RBFs in general case for a better performance-speed tradeoff.
\subsection{Number of Polynomials}
\begin{figure}[ht]
\begin{center}
\centerline{\includegraphics[height=5cm]{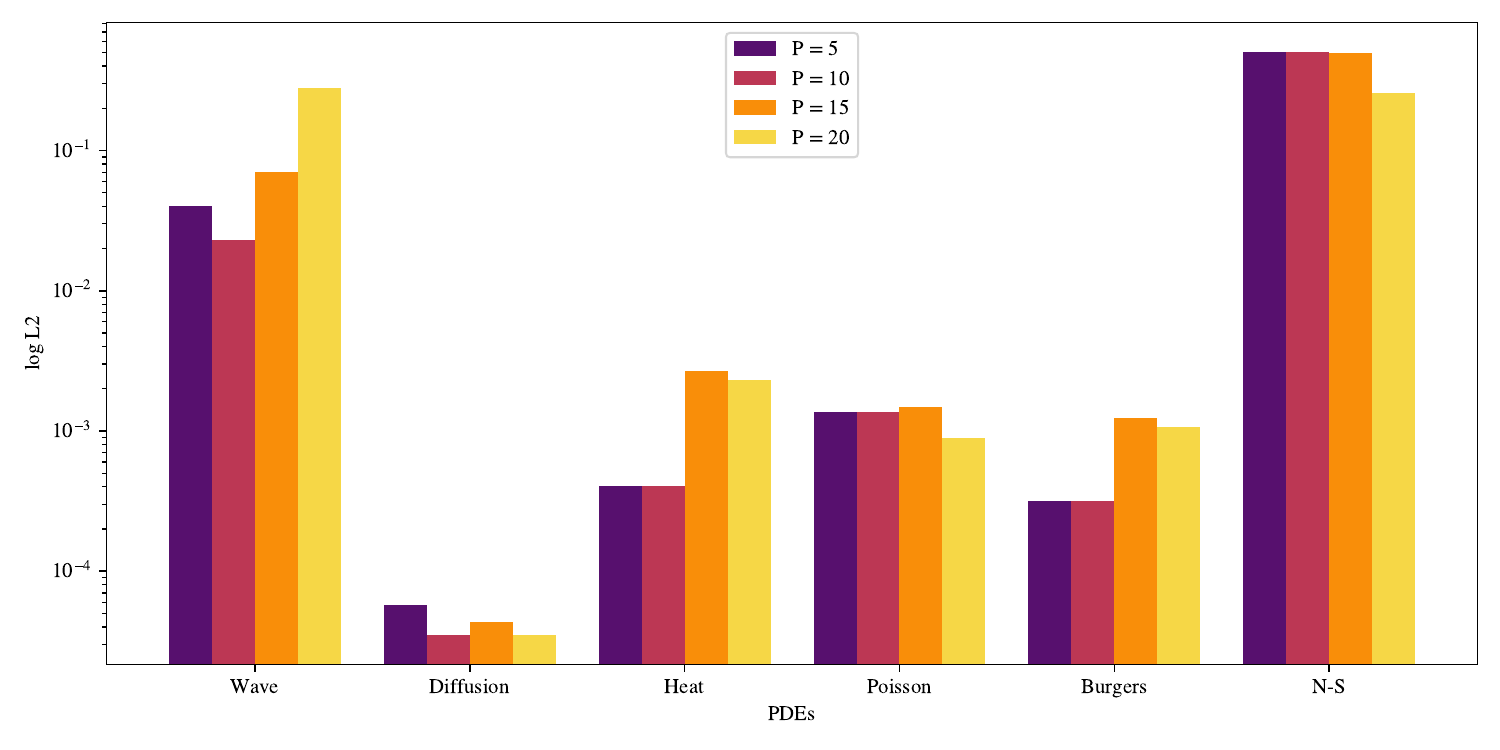}}
\caption{Ablation study on different number of polynomials}
\label{Fig: abla-pol}
\end{center}
\end{figure}

Figure~\ref{Fig: abla-pol} shows an ablation study of how the number of polynomials in feature mappings influences performance in PDEs. It has shown RBF feature mapping with 20 polynomials has achieved best results in the Diffusion equation, Poisson equation and N-S equation. And 10 polynomial terms are better in Heat equation and Burgers equation, thought its performance is matching with only 5 polynomials. 
\subsection{Different Types of RBFs}
Following Table~\ref{tab: RBF types} are common positive definite Radial Basis Functions.
\begin{table}[ht]
\caption{Types of Radial Basis function and their formulation. $\mathbf{x}-\mathbf{c}$ is shorten as r.}
\centering
\label{tab: RBF types}
\begin{tabular}{|l|l|}
\hline
Type                   & Radial function \\ \hline
Cubic                  & $r^3 $              \\ \hline
TPS(Thin Plate Spline) & $r^2log(r)$               \\ \hline
GA(Gaussian)           & $e^{-r^2/\sigma^2}$               \\ \hline
MQ(Multiquadric)       & $\sqrt{1+r^2} $              \\ \hline
IMQ(Inverse MQ)        & $1/\sqrt{1+r^2} $               \\ \hline
\end{tabular}
\end{table}
\begin{figure}[H]
\begin{center}
\centerline{\includegraphics[height=5cm]{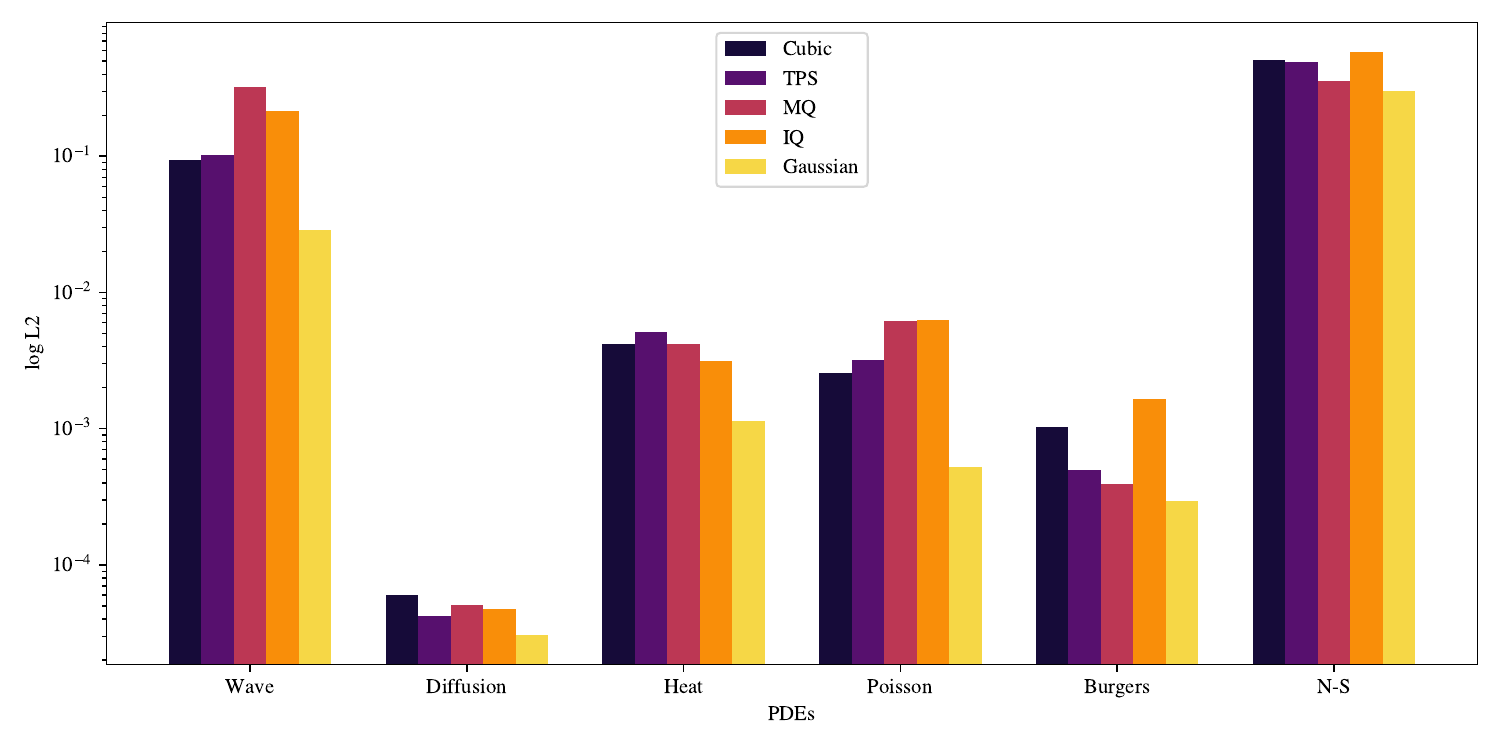}}
\caption{Ablation study on different types of RBFs}
\label{Fig: RBF types}
\end{center}
\end{figure}
The Figure~\ref{Fig: RBF types} has shown Gaussian RBF is dominating all types of PDEs. However other types of RBF are in similar performance. We generally prefer Gaussian RBF in all cases due to its nice properties.

\section{Complexity and Scalability analysis}
 The comparison of complexity and scalability of feature mapping methods are shown in this section.

\label{appendix: complexity}
Although all feature mapping methods are similar in computational complexity, for completeness, we include the complexity of the feature layers that map 128 features and 4 fully connected layers with 50 neurons each. 

\begin{table*}[ht]
\caption{Computational complexity}
\center
\begin{tabular}{|l|l|l|l|l|l|l|l|}
\hline
       & FF     & SF     & RBF-I & RBF-P-5 & RBF-P-10 & RBF-P-15 & RBF-P-20 \\ \hline
FLOPs  & 139.5M & 142.1M & 139.5M  & 142.5M    & 145.0M     & 147.5M     & 150.0M     \\ \hline
Params & 14.2k  & 14.3k  & 14.2k   & 14.5k     & 14.7k      & 14.9k      & 15.2k      \\ \hline
\end{tabular}
\end{table*}

Due to software optimisation and package compatibility, the feature mapping methods can have very different computational efficiency in training.
To demonstrate, we run the above models on different numbers of sample points on Diffusion equation for 3 times in different random seeds.RBF-COM stands for compact support RBF, and RBF-P uses 20 polynomials.\\
The time consumed by Fourier Features is noticeably higher than other methods. All methods have similar runtime for sample points less than $1e4$, that is because all sample points computed are within a GPU parallelisation capacity.
\begin{figure}[ht]
\begin{center}
\centerline{\includegraphics[height=6cm]{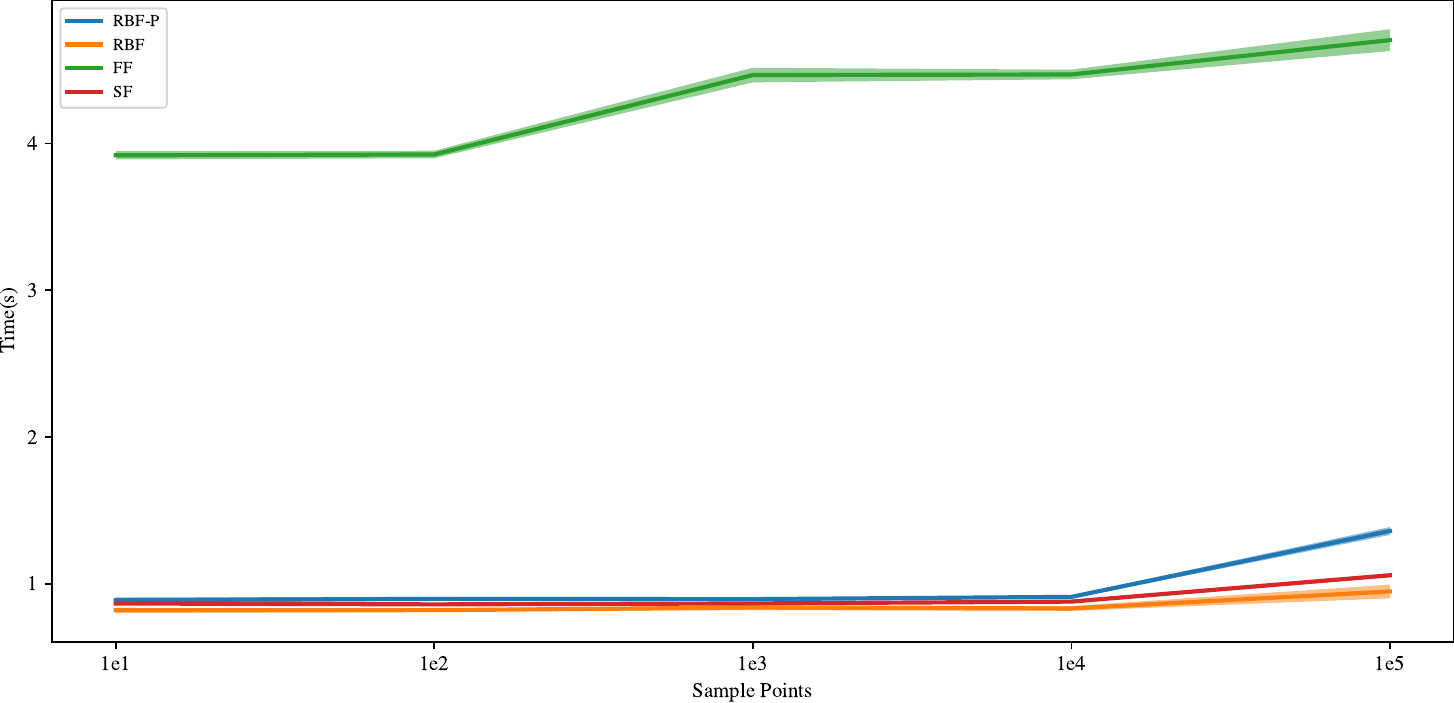}}
\caption{Time consumption on different numbers of sample points with different feature mapping methods}
\label{Fig: complexity}
\end{center}
\end{figure}

\newpage

\section{Benchmark PDEs and Boundary conditions}
\label{Append: Benchmark}

\subsection{Wave Equation}
The one-dimensional Wave Equation is given by:
\begin{equation}
u_{tt} - 4u_{xx}=0.
\end{equation}
In the domain of: 
\begin{equation}
(x, t) \in \Omega \times T = [-1,1] \times[0,1] .
\end{equation}
Boundary condition:
\begin{equation}
u(0, t) = u(1, t) = 0 .
\end{equation}
Initial condition:
\begin{align}
u(x, 0) &= \sin(\pi x) + \frac{1}{2}\sin(4\pi x) \\
u_t &= 0  \\
\end{align}
The analytical solution of the equation is:
\begin{equation}
u(x, t) = \sin(\pi x)cos(2\pi t) + \frac{1}{2}\sin(4\pi x)cos(8\pi t) .
\end{equation}

\subsection{Diffusion Equation}
The one-dimensional Diffusion Equation is given by:
\begin{equation}
u_t - u_{xx} + e^{-t}(sin(\pi x) + \pi^2sin(\pi x))= 0
\end{equation}
In the domain of: 
\begin{equation}
(x, t) \in \Omega \times T = [-1,1] \times[0,1] .
\end{equation}
Boundary condition:
\begin{equation}
u(-1, t) = u(1, t) = 0
\end{equation}
Initial condition:
\begin{align}
u(x, 0) = sin(\pi x)
\end{align}
The analytical solution of the equation is:
\begin{equation}
u(x, t) = e^tsin(\pi x)
\end{equation}
where $\alpha = 0.4, L = 1, n = 1$

\subsection{Heat Equation}
\label{equation: heat}
The two-dimensional Heat Equation is given by:
\begin{equation}
\quad u_{t}-\frac{1}{(500 \pi)^2} u_{x x}-\frac{1}{\pi^2} u_{y y}=0 .
\end{equation}

In the domain of: 
\begin{equation}
(\mathbf{x}, t) \in \Omega \times T = [0,1]^2 \times[0,5] .
\end{equation}

Boundary condition:
\begin{equation}
u(x, y, t)=0 .
\end{equation}

Initial condition:
\begin{align}
u(x, y, 0)=\sin (20 \pi x) \sin (\pi y) .
\end{align}

\subsection{Poisson Equation}
The two-dimensional Poisson Equation is given by:
\begin{equation}
-\Delta u = 0
\end{equation}
In the domain of: 
\begin{equation}
\mathbf{x} \in \Omega = \Omega_{rec} \backslash R_{i}.
\end{equation}
where
\begin{align}
\Omega_{rec} &= [-0.5, 0.5]^2, \\
R_1& =[(x, y):(x-0.3)^2+(y-0.3)^2 \leq 0.1^2], \\
R_2& =[(x, y):(x+0.3)^2+(y-0.3)^2 \leq 0.1^2], \\
R_3& =[(x, y):(x-0.3)^2+(y+0.3)^2 \leq 0.1^2], \\
R_4& =[(x, y):(x+0.3)^2+(y+0.3)^2 \leq 0.1^2].
\end{align}
Boundary condition:
\begin{align}
u &= 0, x \in \partial R_i, \\
u &= 1, x \in \partial \Omega_{rec} .
\end{align}

\subsection{Burgers Equation}
The one-dimensional Burgers Equation is given by:
\begin{equation}
u_t+u u_x=\nu u_{x x}
\end{equation}
In the domain of: 
\begin{equation}
(x, t) \in \Omega=[-1,1] \times[0,1] .
\end{equation}
Boundary condition:
\begin{equation}
u(-1, t)=u(1, t)=0 .
\end{equation}
Initial condition:
\begin{align}
u(x, 0)=-\sin \pi x
\end{align}
where $\nu=\frac{0.01}{\pi}$

\subsection{Steady NS}
The steady incompressible Navier Stokes Equation is given by:
\begin{align}
\nabla \cdot \mathbf{u} & =0 ,\\
\mathbf{u} \cdot \nabla \mathbf{u}+\nabla p-\frac{1}{\operatorname{Re}} \Delta \mathbf{u} & =0. \\
\end{align}

In the domain(back step flow) of: 
\begin{equation}
\mathbf{x} \in \Omega=[0,4] \times[0,2] \backslash\left([0,2] \times[1,2] \cup R_i\right)
\end{equation}
Boundary condition:
\begin{align}
\text{no-slip condition:} \quad \mathbf{u} &= 0 .\\
\text{inlet:} \quad u_x &= 4y(1-y), u_y = 0 .\\
\text{outlet:} \quad p &= 0 .
\end{align}
where $Re = 100$

\subsection{nD Poisson Equation}
The nth-dimensional Poisson Equation is given by:
\begin{equation}
-\Delta u=\frac{\pi^2}{4} \sum_{i=1}^n \sin \left(\frac{\pi}{2} x_i\right)
\end{equation}
In the domain of: 
\begin{equation}
x\in \Omega =[0,1]^n
\end{equation}
Boundary condition:
\begin{equation}
u = 0
\end{equation}
The analytical solution of the equation is:
\begin{equation}
u=\sum_{i=1}^n \sin \left(\frac{\pi}{2} x_i\right)
\end{equation}

\subsection{Inverse Burgers Equation}
The one-dimensional Inverse Burgers Equation is given by:
\begin{equation}
u_t+\mu_1 u u_x=\mu_2 u_{x x}
\end{equation}
In the domain of: 
\begin{equation}
(x, t) \in \Omega=[-1,1] \times[0,1] .
\end{equation}
Boundary condition:
\begin{equation}
u(-1, t)=u(1, t)=0 .
\end{equation}
Initial condition:
\begin{align}
u(x, 0)=-\sin \pi x
\end{align}
where $\mu_1 = 1$ and $\mu_2 =\frac{0.01}{\pi}$

\subsection{Inverse Lorenz Equation}
\label{equation: lorenz}
The 1st-order three-dimensional Lorenz Equation is given by:
\begin{equation}
\begin{aligned}
\frac{dx}{dt} & = \alpha(y - x) ,\\
\frac{dy}{dt} & = x(\rho - z) - y ,\\
\frac{dz}{dt} & = xy - \beta z ,
\end{aligned}
\end{equation}

where $\alpha = 10$, $\beta = \frac{8}{3}$, $\rho = 15$ and the initial points are $x_0 = 0$, $y_0 = 1$, $z_0=1.05$.

\newpage
\section{Visualisations of PDEs solution}
\label{Append: Qualitative}
\begin{figure}[h]
\begin{center}
\begin{tabular}{c} \hspace*{-0.4cm}
\includegraphics[height=20cm]{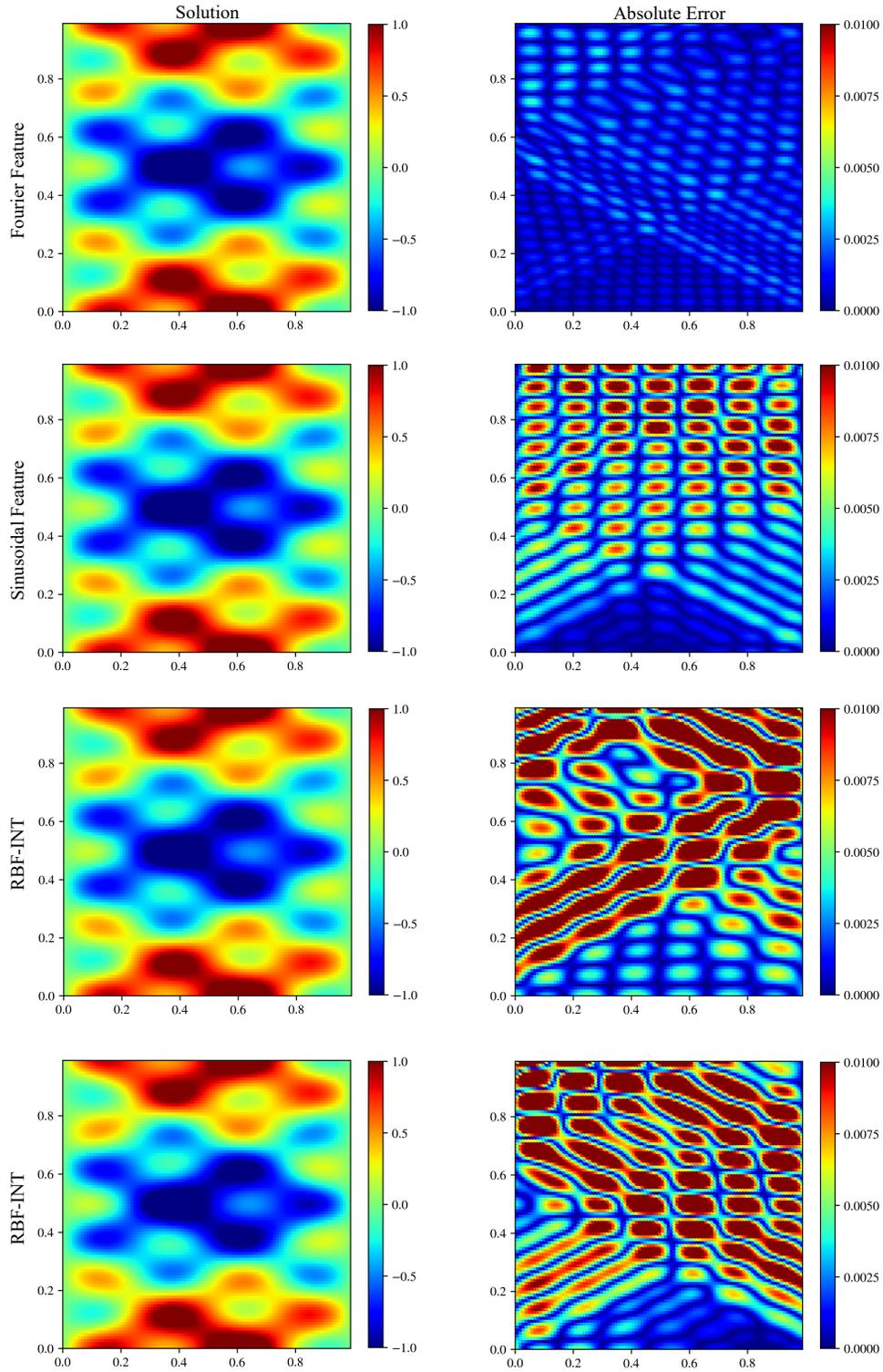}
\end{tabular}
\end{center}
\caption[example] 
{\label{fig:wave qualitative} 
Wave equation}
\end{figure} 

\newpage

\begin{figure*}
\begin{center}
\begin{tabular}{c} \vspace{-30pt}\hspace*{-2cm}
\includegraphics[height=15cm]{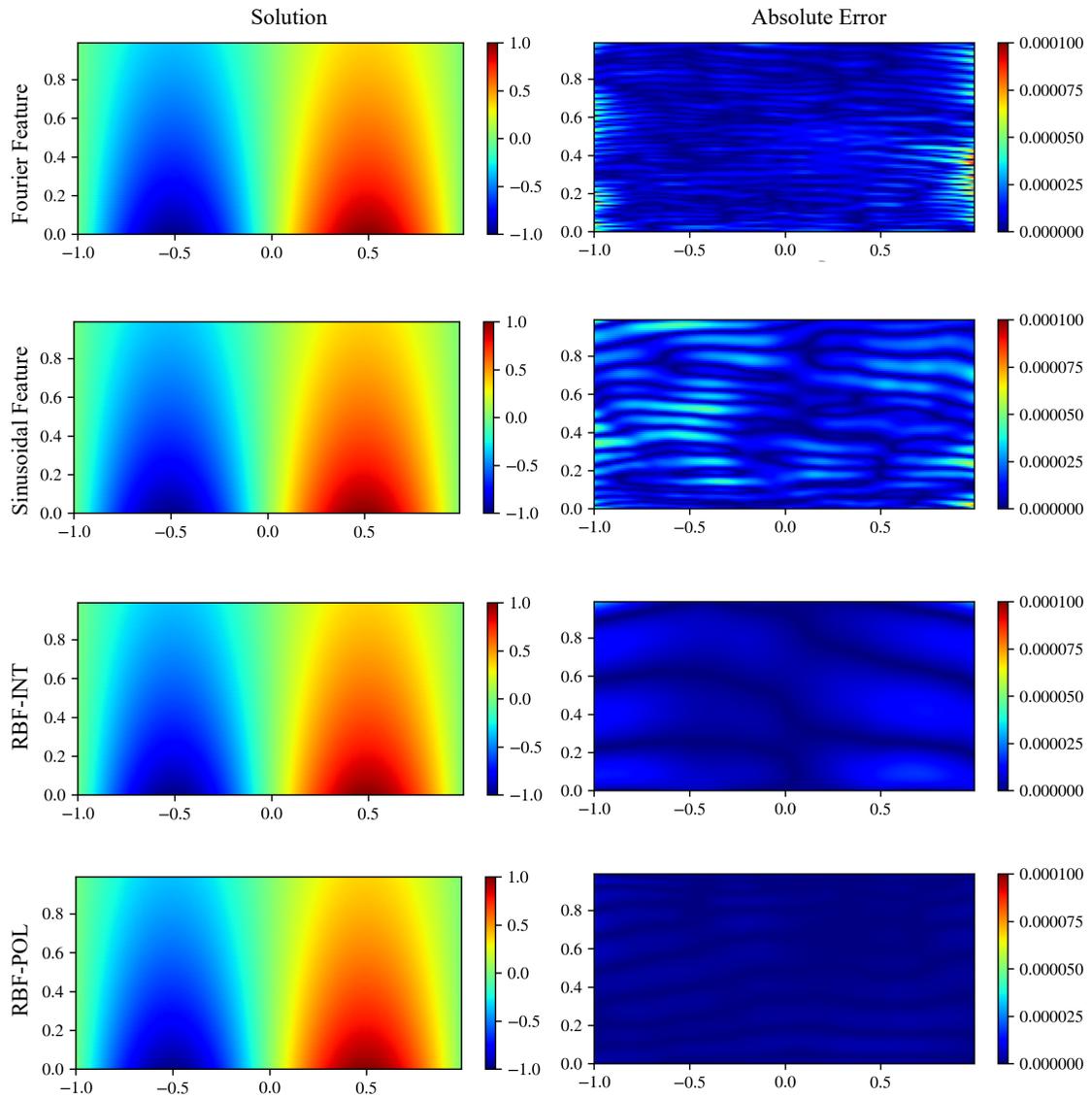}
\end{tabular}
\end{center}
\vspace{20pt}
\caption[example] 
{\label{fig:diffusion qualitative} 
Diffusion equation}
\end{figure*} 

\newpage

\begin{figure*}
\begin{center}
\begin{tabular}{c} \vspace{-30pt}\hspace*{-3cm}
\includegraphics[height=20cm]{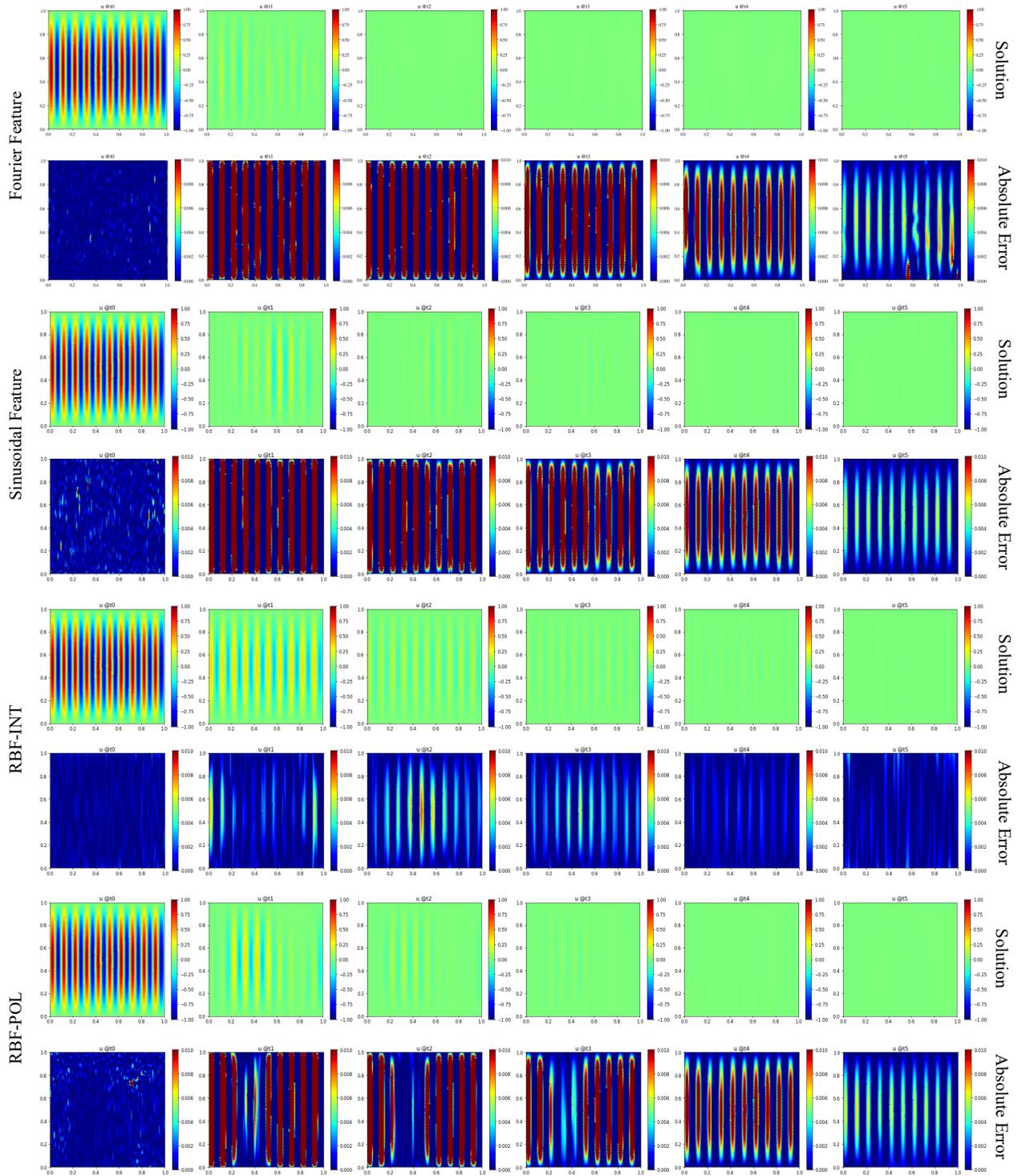}
\end{tabular}
\end{center}
\vspace{20pt}
\caption[example] 
{\label{fig:heat qualitative} 
Heat equation}
\end{figure*}

\newpage

\begin{figure*}
\begin{center}
\begin{tabular}{c} \vspace{-30pt}\hspace*{-2cm}
\includegraphics[height=20cm]{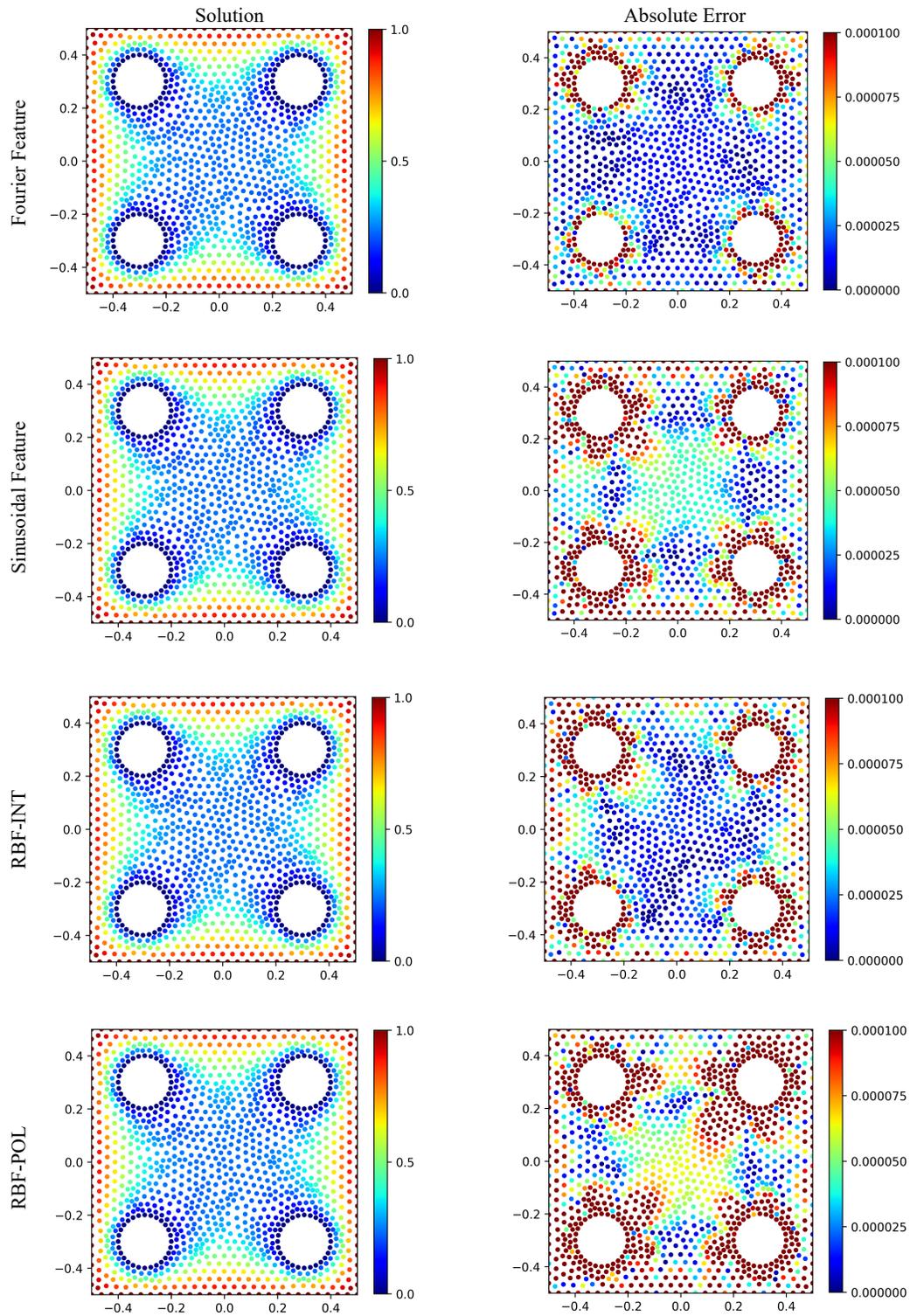}
\end{tabular}
\end{center}
\vspace{20pt}
\caption[example] 
{\label{fig:poisson qualitative} 
Poisson equation}
\end{figure*}

\newpage

\begin{figure*}
\begin{center}
\begin{tabular}{c} \vspace{-30pt}\hspace*{-2cm}
\includegraphics[height=15cm]{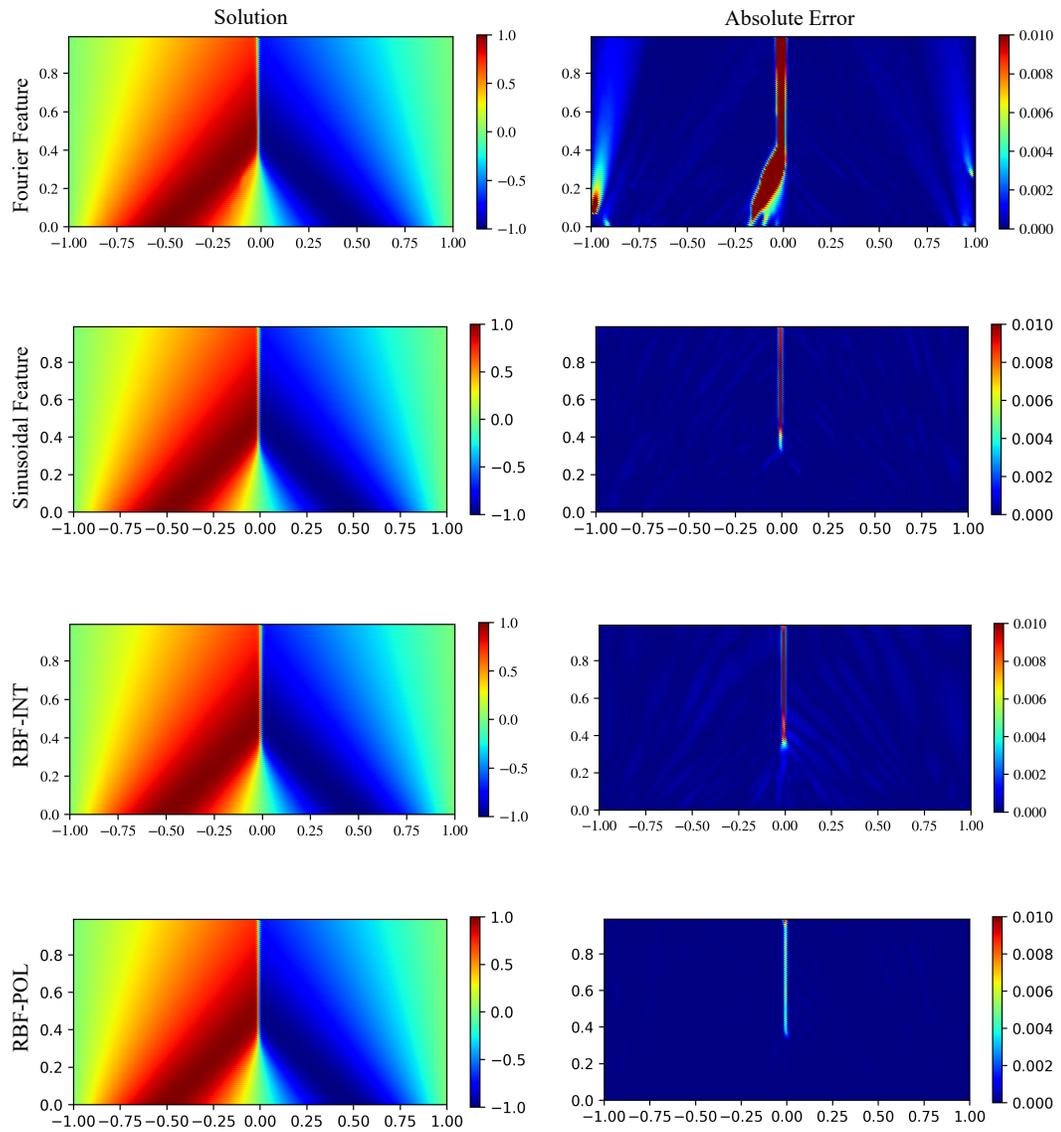}
\end{tabular}
\end{center}
\vspace{20pt}
\caption[example] 
{\label{fig:burgers qualitative} 
 Burgers equation}
\end{figure*} 

\newpage

\begin{figure*}
\begin{center}
\begin{tabular}{c} \vspace{-30pt}\hspace*{-2cm}
\includegraphics[height=20cm]{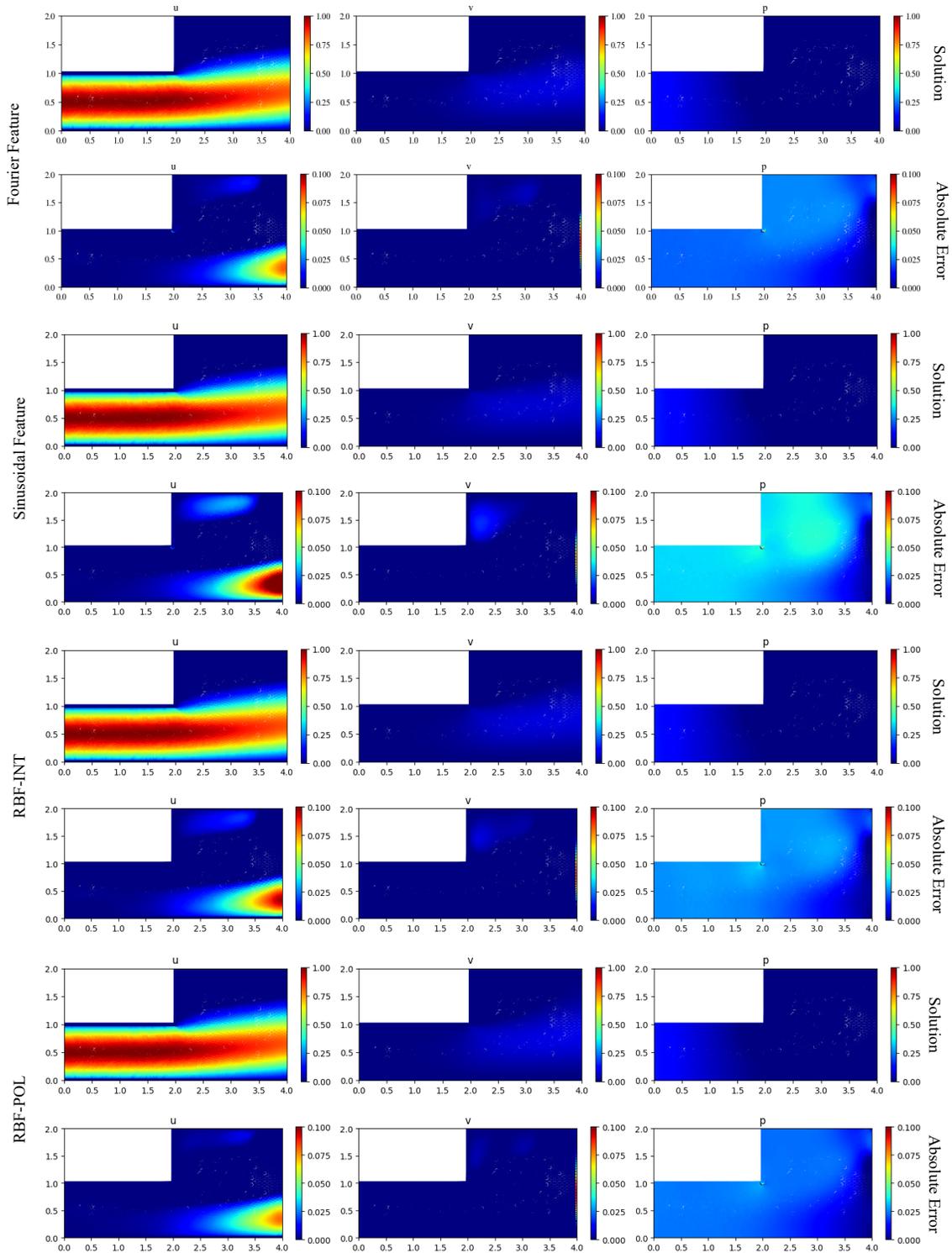}
\end{tabular}
\end{center}
\vspace{20pt}
\caption[example] 
{\label{fig:ns_equation qualitative} 
 Navier-Stokes equation}
\end{figure*}

\end{document}